\documentclass[nohyperref]{article}

\usepackage{microtype}
\usepackage{graphicx}
\usepackage{subfig}
\usepackage{wrapfig}
\usepackage{booktabs} 

\usepackage{threeparttable}
\usepackage[table,xcdraw]{xcolor}
\usepackage{multirow}
\usepackage{float}
\usepackage{bm}

\usepackage{amsmath,amsfonts,bm}

\def\eqref#1{equation~\ref{#1}}

\def\1{\bm{1}}

\def\vtheta{{\bm{\theta}}}

\def\vh{{\bm{h}}}

\def\vu{{\bm{u}}}

\def\vx{{\bm{x}}}
\def\vy{{\bm{y}}}

\DeclareMathAlphabet{\mathsfit}{\encodingdefault}{\sfdefault}{m}{sl}
\SetMathAlphabet{\mathsfit}{bold}{\encodingdefault}{\sfdefault}{bx}{n}

\DeclareMathOperator*{\argmin}{arg\,min}

\usepackage{hyperref}

\usepackage[accepted]{icml2023}

\usepackage{amsmath}
\usepackage{amssymb}
\usepackage{mathtools}
\usepackage{amsthm}

\usepackage[capitalize,noabbrev]{cleveref}

\theoremstyle{plain}
\newtheorem{theorem}{Theorem}[section]

\theoremstyle{definition}

\theoremstyle{remark}

\usepackage[textsize=tiny]{todonotes}

\def\ddefloop#1{\ifx\ddefloop#1\else\ddef{#1}\expandafter\ddefloop\fi}

\def\ddef#1{\expandafter\def\csname v#1\endcsname{\ensuremath{\boldsymbol{#1}}}}
\ddefloop ABCDEFGHIJKLMNOPQRSTUVWXYZabcdefghijklmnopqrstuvwxyz\ddefloop

\def\ddef#1{\expandafter\def\csname v#1\endcsname{\ensuremath{\boldsymbol{\csname #1\endcsname}}}}
\ddefloop {alpha}{beta}{gamma}{delta}{epsilon}{varepsilon}{zeta}{eta}{theta}{vartheta}{iota}{kappa}{lambda}{mu}{nu}{xi}{pi}{varpi}{rho}{varrho}{sigma}{varsigma}{tau}{upsilon}{phi}{varphi}{chi}{psi}{omega}{Gamma}{Delta}{Theta}{Lambda}{Xi}{Pi}{Sigma}{varSigma}{Upsilon}{Phi}{Psi}{Omega}{ell}\ddefloop

\def\ddef#1{\expandafter\def\csname bb#1\endcsname{\ensuremath{\mathbb{#1}}}}
\ddefloop ABCDEFGHIJKLMNOPQRSTUVWXYZ\ddefloop

\icmltitlerunning{Continuation Path Learning for Homotopy Optimization}

\begin{document}

\twocolumn[
\icmltitle{Continuation Path Learning for Homotopy Optimization}




\begin{icmlauthorlist}
\icmlauthor{Xi Lin}{cityu}
\icmlauthor{Zhiyuan Yang}{cityu}
\icmlauthor{Xiaoyuan Zhang}{cityu}
\icmlauthor{Qingfu Zhang}{cityu}
\end{icmlauthorlist}

\icmlaffiliation{cityu}{Department of Computer Science, City University of Hong Kong}

\icmlcorrespondingauthor{Xi Lin}{xi.lin@my.cityu.edu.hk}

\icmlkeywords{Homotopy Optimization, Continuation Optimization, Learning to Optimize, Continuation Path Learning}

\vskip 0.3in
]



\printAffiliationsAndNotice{}  

\begin{abstract}

Homotopy optimization is a traditional method to deal with a complicated optimization problem by solving a sequence of easy-to-hard surrogate subproblems. However, this method can be very sensitive to the continuation schedule design and might lead to a suboptimal solution to the original problem. In addition, the intermediate solutions, often ignored by classic homotopy optimization, could be useful for many real-world applications. In this work, we propose a novel model-based approach to learn the whole continuation path for homotopy optimization, which contains infinite intermediate solutions for any surrogate subproblems. Rather than the classic unidirectional easy-to-hard optimization, our method can simultaneously optimize the original problem and all surrogate subproblems in a collaborative manner. The proposed model also supports real-time generation of any intermediate solution, which could be desirable for many applications. Experimental studies on different problems show that our proposed method can significantly improve the performance of homotopy optimization and provide extra helpful information to support better decision-making.

\end{abstract}

\section{Introduction}
\label{sec_intro}

Homotopy optimization~\cite{blake1987visual, yuille1989energy,allgower1990numerical}, also called continuation optimization, is a general optimization strategy for solving complicated and highly non-convex optimization problems which can be found in many machine learning applications~\cite{jain2017non}. This method first constructs a  simple surrogate of the original optimization problem, and then gradually solves a sequence of easy-to-hard surrogates to approach the optimal solution of the original complicated problem. The simplest surrogate subproblem could be easily solved, and its solution will serve as a good initial one for the next subproblem. In this way, we can eventually find a good initial and then a (nearly) optimal solution for the original hard-to-solve optimization problem.   

The idea of homotopy optimization is straightforward but it also suffers several drawbacks. First, the optimization performance could heavily depend on the continuation schedule of the surrogate subproblems, which is not easy to design for a new problem~\cite{dunlavy2005homotopy}. It also needs to iteratively solve each subproblem in sequence, which could lead to undesirable long run time in practice~\cite{iwakiri2022single}. In addition, the existing homotopy optimization methods only focus on finding the final solution to the original problem. However, the intermediate solutions for homotopy surrogate subproblems could be useful for many real-world applications.    

In this work, to tackle the drawbacks mentioned above, we propose a novel continuation path learning (CPL) method to find the whole solution path for homotopy optimization, which contains infinite solutions for all intermediate subproblems. The key idea is to build a learnable model that maps any valid continuation level to its corresponding solution, and then optimize all of them simultaneously in a collaborative manner. In this way, the whole path model can be learned in a single run without sensitive schedule design and iterative optimization. With the learned path model, we can easily generate the solution for any intermediate homotopy level, which could be useful for many applications. Our main contributions can be summarized as follows:   

\begin{figure*}[t]
    \centering
    \includegraphics[width= \linewidth]{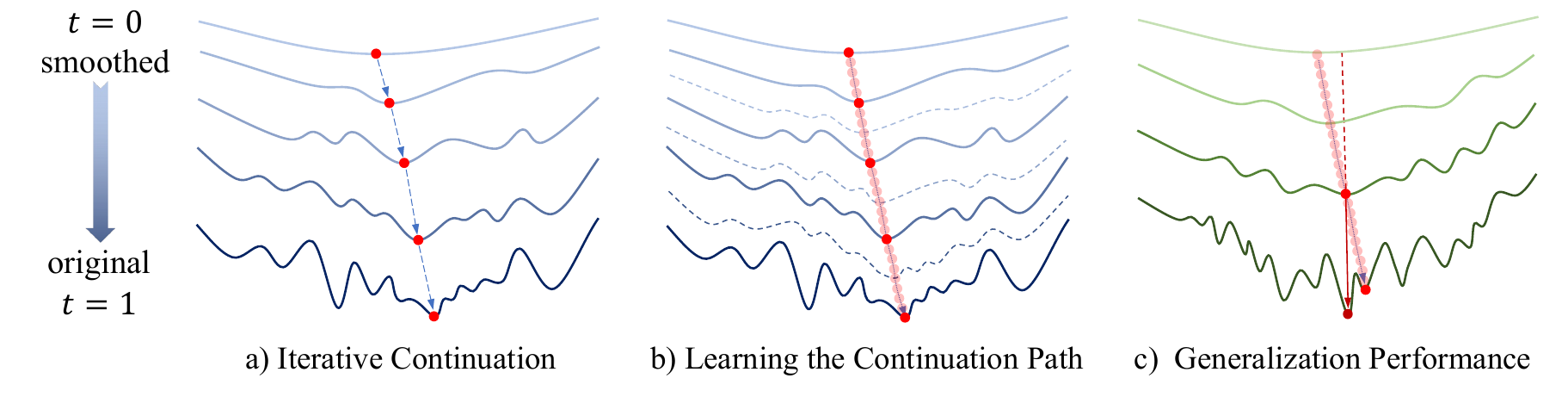}
    \caption{\textbf{Continuation Path Learning}: \textbf{a)} The classical homotopy optimization method sequentially solves a set of easy-to-hard smoothed subproblems, which helps find the optimal solution for the original problem. \textbf{b)} We propose to simultaneously learn the whole continuation path, which contains the intermediate solutions for all homotopy subproblems. \textbf{c)} The solutions for homotopy subproblems could have better generalization performance for learning-based problems.}
    \label{fig_continuation_idea}
    \vspace{-0.2in}
\end{figure*}

\begin{itemize}
    \item We propose a novel model-based approach to learn the whole continuation path for homotopy optimization, which is significantly different from the existing methods that iteratively solve a sequence of finite subproblems. 
    
    \item We develop an efficient learning method to train the path model concerning all homotopy levels simultaneously. The proposed model can generate solutions for any intermediate subproblem in real time, which is desirable for many real-world applications.  
    
    \item We empirically demonstrate that our proposed CPL method can achieve promising performances on various problems, including non-convex optimization, noisy regression, and neural combinatorial optimization.~\footnote{The source code can be found in \url{https://github.com/Xi-L/CPL}.}
\end{itemize}

\section{Related Work}
\label{sec_related_work}

\textbf{Homotopy Optimization}, also called continuation or graduated optimization method~\cite{blake1987visual, yuille1989energy,allgower1990numerical}, is a general optimization strategy for solving non-convex optimization problems. This method sequentially constructs and solves a set of smoothed subproblems that gradually deform from an easy-to-solve problem to the original complicated problem as shown in Figure~\ref{fig_continuation_idea}(a). It would help to find a better solution for the original non-convex problem~\cite{wu1996effective, dunlavy2005homotopy}. This method has also been widely used for solving nonlinear equations~\cite{eaves1972homotopies,wasserstrom1973numerical,allgower1990numerical}, and it is closely related to simulated annealing for optimization~\cite{kirkpatrick1983optimization, van1987simulated, ingber1993simulated}.

\paragraph{Homotopy Optimization in Machine Learning} The homotopy optimization methods have been widely used in different machine learning applications over the past three decades, such as for computer vision~\cite{terzopoulos1988computation, gold1994learning, brox2010large, hruby2022learning}, statistical learning~\cite{chapelle2006continuation,kim2010gaussian}, curriculum learning~\cite{bengio2009learning, bengio2009curriculum,kumar2010self,graves2017automated}, and efficient model training~\cite{chaudhari2016entropy, gargiani2020transferring,guo2020batch}. A few works have been proposed to study its theoretical property in different settings~\cite{mobahi2015theoretical, mobahi2015link, hazan2016graduated, anandkumar2017homotopy, iwakiri2022single}. Although the homotopy optimization method can usually help to find a better solution, it might suffer from a long run time due to the iterative optimization structure. Recently, \citet{iwakiri2022single} have proposed a novel single loop algorithm for fast Gaussian homotopy optimization.  

Most homotopy optimization methods only care about the final solution for the original problem, but the intermediate optimal solutions could also be useful for flexible decision-making. A few works have been proposed to find a single solution for a specific smooth intermediate subproblem for better generalization performance~\cite{chaudhari2016entropy,gulcehre2016mollifying}. In statistical learning, efficient methods have been proposed to find the whole set of finite tuning points that fully characterize the homotopy path for LASSO~\cite{efron2004least,rosset2007piecewise,tibshirani2011solution} and SVM~\cite{hastie2004entire} by leveraging the specific piece-wise linear structure. However, these methods do not work for general homotopy optimization problems. 

\paragraph{Model-based Optimization} Many model-based methods have been proposed to improve different optimization algorithms' performance. Bayesian optimization builds a surrogate model to approximate the unknown black-box optimization problem and uses it to guide the optimization process~\cite{shahriari2016taking,garnett2022bayesian}. Latent space modeling~\citep{gomez2018automatic, tripp2020sample} is another powerful approach for reconstructing the original complicated optimization problem into a much easier form to solve. It is also possible to accelerate the optimization algorithm by learning the problem structure~\cite{sener2020learning} or dividing the search space~\cite{wang2020learning, eriksson2019scalable}. Recently, a few model-based approaches have been proposed to learn the Pareto set for multi-objective optimization problems~\cite{yang2019generalized, dosovitskiy2019you, lin2020controllable, navon2021learning, lin2022pareto_combinatorial, lin2022pareto_expensive}. In this work, we propose a novel method to learn the whole continuation path for homotopy optimization as shown in Figure~\ref{fig_continuation_idea}(b), and use it to improve the optimization performance for different applications.

\clearpage

\section{Homotopy Optimization}
\label{sec_homotopy_opt}

In this section, we introduce the classical homotopy optimization method and a recently proposed single-loop Gaussian homotopy algorithm.

\subsection{Classical Homotopy Optimization}

We are interested in the following minimization problem:
\begin{eqnarray}\label{min_problem}
\min_{\vx \in \mathcal{X}} f(\vx),
\end{eqnarray} 
where $\vx \in \mathcal{X} \subset \bbR^d$ is the decision variable, and $f:\mathcal{X} \rightarrow \bbR$ is the objective function to minimize. The objective function $f(\vx)$ could be highly non-convex and has a complicated optimization landscape. Therefore, it cannot be easily optimized with a simple local minimization method such as gradient descent. 

To tackle this problem, homotopy optimization method~\cite{allgower1990numerical,dunlavy2005homotopy} considers a family of function $H: \mathcal{X} \times \mathcal{T} \rightarrow \bbR$ parameterized by the continuation level $t \in \mathcal{T} = [0,1]$ such that:
\begin{align}
H(\vx,t = 0) = g(\vx) \quad H(\vx, t = 1) = f(\vx), \forall \vx \in \mathcal{X},
\end{align}
where $g: \mathcal{X} \rightarrow \bbR$ is another easy-to-optimize objective function on the same decision space $\mathcal{X}$. The function $H(\vx,t)$ is also called a homotopy that gradually transforms $g(\vx)$ to $f(\vx)$ by increasing $t$ from $0$ to $1$. An illustration of the continuation function can be found in Figure~\ref{fig_continuation_idea}(a).

The key idea of homotopy optimization is to define a suitable continuation function $H(\vx,t)$ such that the minimizer for $H(\vx,0) = g(\vx)$ is already known or easy to find, and the $H(\vx,t)$ with $t = 0 \rightarrow 1$ be a sequence of smoothed functions transforming from $g(\vx)$ to the target objective function $f(\vx)$. Rather than directly optimizing the complicated target function $f(\vx)$, we can progressively solve a sequence of coarse-to-fine smoothed optimization subproblems from $H(\vx,0)$ to $H(\vx,1)$ with a warm start from previously obtained solution as shown in \textbf{Algorithm~\ref{alg_continuaiton}}. In this way, we can find a better solution for the target objective function $H(\vx,1) = f(\vx)$. 

In practice, the performance of homotopy optimization heavily depends on two crucial components:
\begin{itemize}
    \item A proper design of the initial and continuation function for the given problem;
    \item A suitable schedule to progressively solve the sequence of easy-to-hard subproblems.
\end{itemize}
However, there is no clear and principled guideline for the continuation construction~\cite{mobahi2015link}. The homotopy optimization process could be time-consuming since we have to run a local search algorithm to find $\vx_k$ for each subproblem~\cite{iwakiri2022single}. The continuation and schedule design will become much more complicated if we want to find a solution for a specific homotopy level unknown in advance, such as for better generalization~\cite{chaudhari2016entropy,gulcehre2016mollifying}.  

\begin{algorithm}[t]
\caption{Classical Homotopy Optimization Algorithm}
\label{alg_continuaiton}
    \begin{algorithmic}[1]
    \STATE \textbf{Input:} continuation function $H(\vx, t)$, a predefined sequence $0 = t_0 < t_1 < \ldots t_K = 1$
    \STATE $\vx_0 =$ a minimizer of $H(\vx,t_0)$ 
    \FOR{$k = 1$ to $K$}
        \STATE $\vx_k =$ local minimizer of $H(\vx,t_k)$, initialized at $\vx_{k-1}$
    \ENDFOR
    \STATE{Output:} $\vx_K$
    \end{algorithmic}
\end{algorithm}

\subsection{Single Loop Gaussian Homotopy Algorithm}

\begin{algorithm}[H]
\caption{Single Loop Gaussian Homotopy Algorithm}
\label{alg_slgh}
    \begin{algorithmic}[1]
    \STATE \textbf{Input:} Gaussian homotopy function $GH(\vx, t)$, initial solution $\vx_0$, initial homotopy level $t_0$
    \FOR{$k = 1$ to $K$}
        \STATE $\vx_k = \vx_{k-1} - \eta_1  \nabla_{\vx} GH(\vx_{k-1}, t_{k-1})$ 
        \STATE (optional) query $G_t = \frac{\partial GH(\vx_{k-1}, t_{k-1})}{\partial t}$
        \STATE update $t_k = $
            \begin{align}
            \begin{split}
            \left \{
            \begin{array}{ll}
                \gamma t_{k-1}&(\text{SLGH}_r)\\
                \max\{0, \min\{t_{k-1} - \eta_2 G_t, \gamma t_{k-1}\}\} &(\text{SLGH}_d) \nonumber
            \end{array}
            \right.
            \end{split}
            \end{align}
    \ENDFOR
    \STATE{Output:} $\vx_K$
    \end{algorithmic}
\end{algorithm}

To alleviate the time-consuming local optimization at each homotopy iteration, \citet{iwakiri2022single} recently proposed a novel single loop algorithm for the popular Gaussian homotopy method~\cite{blake1987visual}.

The Gaussian homotopy function $GH(\vx,t)$ with $t \in [0,1]$ for $f(\vx)$ can be defined as:
\begin{eqnarray}\label{gaussian_homotopy}
\begin{aligned}
   GH(\vx,t) &= \bbE_{\vu \sim N(0,I_d)}[f(\vx + \beta(1-t)\vu)] \\
    &= \int f(\vx + \beta(1-t)\vy)k(\vy)d\vy,
\end{aligned} 
\end{eqnarray}
where parameter $\beta > 0$ controls the max range of homotopy effect, $N(0,I_d)$ is the $d$-dimensional standard Gaussian distribution, and $k(\vy) = (2\pi)^{-d/2}\exp(-||\vy||^2/2)$ is the Gaussian kernel. Instead of iteratively optimizing $GH(\vx,t)$ with a sequence of $t$ as in previous works~\cite{wu1996effective, mobahi2015theoretical, mobahi2015link, hazan2016graduated}, \citet{iwakiri2022single} proposed to directly optimize:   
\begin{align}\label{single_loop_gaussian_homotopy}
\min_{\vx \in \mathcal{X}, t \in \mathcal{T}} GH(\vx,t)
\end{align} 
with respect to both $\vx$ and $t$ at the same time. By leveraging the theoretical properties of the heat equation~\cite{widder1976heat} and partial differential equation~\cite{evans2010partial}, they also showed that the Gaussian homotopy function $GH(\vx,t)$ will be always optimized at $(\vx^*,1)$ where $\vx^*$ is the optimal solution for $f(\vx)$. This property validates the approach to optimize $GH(\vx,t)$ with respect to $t$.

The single loop Gaussian homotopy (SLGH) optimization algorithm is shown in \textbf{Algorithm~\ref{alg_slgh}}. At each iteration, the decision variable $\vx_k$ is updated by gradient descent, and the homotopy level $t_k$ is updated by either a fixed ratio decreasing rule (e.g., $\text{SLGH}_r$ with $\gamma = 0.999$) or a derivative update rule (e.g., $\text{SLGH}_d$). This algorithm could be faster than the classical homotopy optimization algorithm with a double loop structure~\cite{iwakiri2022single}. However, the performance of SLGH is still sensitive to the homotopy schedule (e.g., the setting of $\gamma$). It might not work for a general homotopy optimization problem other than Gaussian homotopy, and can not easily obtain the solutions for any intermediate subproblem or the whole homotopy path. 

\section{Continuation Path Learning}
\label{sec_cpl}

\subsection{Why Continuation Path Learning}

Both classical homotopy optimization algorithms and the recently proposed SLGH method only focus on finding a single solution for the original optimization problem. In many real-world applications, an intermediate solution $\vx^{t^*}$ for a smooth homotopy subproblem $H(\vx^{t^*},t^*)$ could be desirable such for robust generalization performance~\cite{chaudhari2016entropy,gulcehre2016mollifying}.Finding a solution with a proper regularization v.s. performance trade-off is also an important issue in machine learning~\cite{efron2004least,hastie2004entire}. However, the optimal homotopy level $t^*$ is usually unknown in advance.     

In this work, we propose a novel model-based approach to learn the whole continuation path that contains solutions for all gradually smooth homotopy subproblems. With the learned path model, decision-makers can easily select their preferred solution(s) on the solution path as shown in Figure~\ref{fig_continuation_path_model}. In addition, even if the goal is to find a single solution with a given homotopy level (e.g., $t=1$ for the original problem), CPL can also efficiently generate a good initial solution by collaboratively learning all subproblems together. It is a strong alternative to the sequential and unidirectional information passing in classical homotopy optimization.  

\begin{figure}[t]
    \centering
    \includegraphics[width= \linewidth]{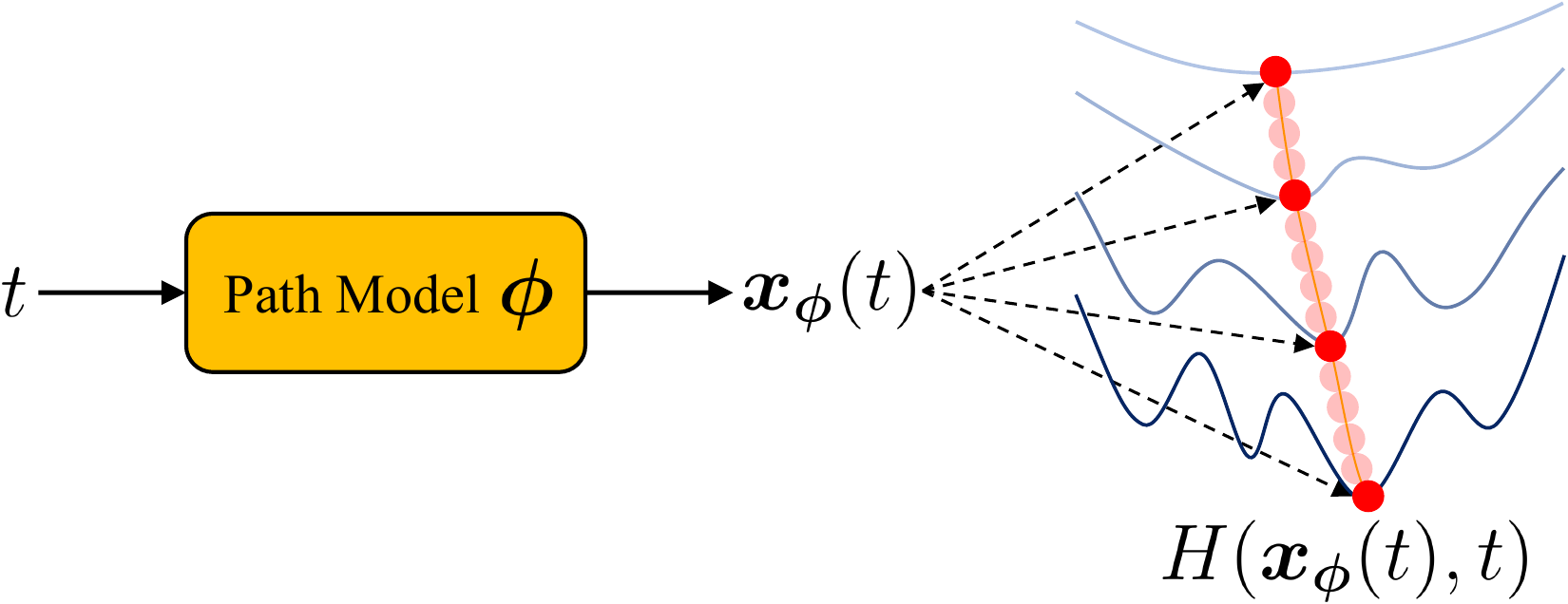}
    \caption{\textbf{Continuation Path Model} takes any valid homotopy level $t$ as input, and generate its corresponding solution $\vx_{\vphi}(t)$ on the continuation path. Decision-makers can easily obtain any predictive homotopy solution by adjusting the input level $t$.}
    \label{fig_continuation_path_model}
\end{figure}

\subsection{Continuation Path Model}

For the homotopy function $H(\vx,t)$, the number of continuation levels $t \in \mathcal{T}$ and their corresponding subproblems could be infinite. Let $\vx_0^*$ be the minimizer for $H(\vx,0) = g(\vx)$, we can define a path $\vx^*(t)$ continuous in $t$ such that $\vx^*(0) = \vx_0^*$ and $\nabla H(\vx^*(t),t) = 0$ for all $t \in \mathcal{T} = [0,1]$~\cite{wu1996effective}. This path simply goes through a set of stationary points for $H(\vx,t)$ from $\vx_0^*$ with gradually increasing $t$. The homotopy optimization algorithms trace the solutions from $\vx^*(0)$ to $\vx^*(1)$. In practice, the solution $\vx^*(1)$ could be a good local minimizer for $H(\vx,1) = f(\vx)$ if not the global one~\cite{mobahi2015link, mobahi2015theoretical}. The existence and uniqueness of this path can be guaranteed for the Gaussian homotopy function (\ref{gaussian_homotopy}) under mild conditions: 
\begin{theorem}[Existence of Continuation Path~\citep{wu1996effective}]
Let $f$ be a well-behaved function and $H(\vx, t)$ be its Gaussian homotopy function (\ref{gaussian_homotopy}). Then for any stationary point $x_0$ of $H(\vx,0)$, there is a continuous and differentiable curve $\vx^*(t)$ on $t \in \mathcal{T} = [0,1]$ such that $\vx^*(0) = x_0$ and $x^*(t)$ is a stationary solution of $H(\vx,t), \forall t \in \mathcal{T}$.
\end{theorem}

To be well-behaved, a sufficient condition is that the function $f$ is twice continuously differentiable while $f$ and its derivatives should all be integrable for the Gaussian homotopy function (\ref{gaussian_homotopy})~\cite{wu1996effective}. We assume such continuation path $\vx^*(t)$ always exists in this work.   

According to the definition, $\vx^*(t)$ is a continuous curve that contains solutions for all (infinite) homotopy levels $t \in \mathcal{T}$. The discrete set of stationary solutions obtained by a classical homotopy optimization $\{\vx_1,\vx_2,\ldots, \vx_K\} = \{\vx^*(t_1),\vx^*(t_2),\ldots, \vx^*(t_K)\}$ is a finite subset on the solution path $\{\vx^*(t)|t \in \mathcal{T}\}$. In this work, we propose to build a model $\vx_{\vphi}(t)$ with learnable parameter $\vphi$ to approximate the whole continuation path $\vx^*(t)$. Our goal is to find the optimal $\vphi^*$ such that:
\begin{eqnarray}\label{continuation_path}
\vx_{\vphi^*}(t) = \vx^*(t) = \argmin_{\vx} H(\vx,t), \forall t \in \mathcal{T}.
\end{eqnarray} 
As shown in Figure~\ref{fig_continuation_path_model}, the continuation path model maps any valid homotopy level $t$ to its corresponding solution $\vx_{\vphi}(t)$. With the ideal model parameters $\vphi^*$, the output $\vx_{\vphi^*}(t)$ should be the optimal solutions for each intermediate subproblem $H(\vx,t)$, and hence $\vx_{\vphi^*}(t)$ well approximates the continuation path $\vx^*(t)$. Once such a model is obtained, we can easily get the corresponding solution $\vx_{\phi^*}(t^\prime) = \argmin_{\vx} H(\vx,t^\prime)$ for any specific continuation parameter $t^\prime$. In this work, we set $\vx_{\phi}(t)$ as a neural network model and $\vphi$ is its model parameters.

\subsection{Learning the Continuation Path}
\label{subsec_path_learning}

Once we have the continuation path model $\vx_{\vphi}(t)$, the next step is to find the optimal parameters $\vphi^*$ with respect to all homotopy level $t \in \mathcal{T}$. Since the number of homotopy levels is infinite, our goal is to optimize the following expectation:   
\begin{align} \label{eq_homotopy_expectation}
\min_{\vphi} \bbE_{t} H(\vx_{\vphi}(t),t),
\end{align}
where each term $H(\vx_{\vphi}(t),t)$ is a composition of continuation path model $\vx_{\vphi}(t)$ and the homotopy function $H(\vx,t)$. In this way, we reformulate the classic unidirectional homotopy optimization problem (e.g., \textbf{Algorithm~\ref{alg_continuaiton}}) into the single loop model training problem~(\ref{eq_homotopy_expectation}) that simultaneously learn the whole continuation path. \textit{It should be noticed that our method changes the optimization variables from the original decision variable $\vx$ to model parameter $\vphi$.} 

It is difficult to directly optimize the problem~(\ref{eq_homotopy_expectation}) since the expectation term could be hard to compute in most cases. In this work, we propose to learn the model parameters with stochastic gradient descent as shown in \textbf{Algorithm~\ref{alg_path_learning}}. At each step, we optimize the following stochastic optimization problem with Monte Carlo sampling:  
\begin{align} \label{eq_homotopy_mc_sampling}
\min_{\vphi} \frac{1}{M} \sum_{m=1}^{M}  H(\vx_{\vphi}(t_m),t_m), \quad \{t_m\}_{m=1}^{M} \sim P_{\mathcal{T}},
\end{align}
where $\{t_1,\ldots,t_M\}$ are $M$ independent identically distributed (i.i.d.) samples from distribution $P_{\mathcal{T}}$. Without any prior knowledge, we can simply set $P_{\mathcal{T}}$ to be a uniform distribution on $\mathcal{T}$. It is also possible to use other distributions or further adaptively adjust the distribution to incorporate extra information along the optimization process.

A crucial step of the proposed method is to find a valid gradient direction to update the model parameters at each iteration. We can decompose the gradient $\nabla_{\vphi} H(\vx_{\vphi}(t), t)$ with the chain rule:
\begin{align} \label{eq_homotopy_gradient_chain_rule}
\nabla_{\vphi} H(\vx_{\vphi}(t), t) = \frac{\partial \vx_{\vphi}(t) }{\partial \vphi} \cdot \nabla_{\vx} H(\vx = \vx_{\vphi}(t),t),
\end{align}
where $\frac{\partial \vx_{\vphi}(t) }{\partial \vphi}$ is the Jacobian matrix of the path model with output vector $\vx_{\vphi}(t)$, and $\nabla_{\vx} H(\vx,t)$ is the gradient of the homotopy function with respect to decision variables $\vx$. In this work, since the path model is a neural network, the Jacobian matrix $\frac{\partial \vx_{\vphi}(t) }{\partial \vphi}$ can be easily calculated with backpropagation. If the homotopy function is also differentiable with a known gradient formulation $\nabla_{\vx} H(\vx,t)$, we can use standard gradient descent to optimize the model parameters. 

\begin{algorithm}[t]
\caption{Gradient-based Continuation Path Learning}
    \label{alg_path_learning}
    \begin{algorithmic}[1]
    \STATE \textbf{Input:} continuation function $H(\vx, t)$, a path model $\vx_{\phi}(t)$ with learnable parameters $\vphi$
    \FOR{$i = 1$ to $I$}
        \STATE randomly sample a set of $\{t_m\}_{m=1}^{M} \sim P_{\mathcal{T}}$
        \STATE $\vphi \leftarrow \vphi - \frac{\eta}{M} \sum_{m=1}^{M} \nabla_{\vphi} H(\vx = \vx_{\vphi}(t_m), t_m)$
    \ENDFOR
    \STATE (optional) $\vx^{t^{\prime}} =$ local minimizer of $H(\vx,t^{\prime})$, initialized at $\vx_{\vphi}(t^{\prime})$ with chosen homotopy level $t^{\prime}$
    \STATE{Output:} path model $\vx_{\vphi}(t)$ 
    \end{algorithmic}
\end{algorithm}

In many real-world applications, however, the gradient of the homotopy function could be unknown or hard to compute~\cite{iwakiri2022single}. In these cases, we can use a zeroth-order optimization (also called derivative-free optimization) method~\cite{duchi2015optimal,nesterov2017random} with approximate gradients for model training. For a general homotopy function, we can adopt a simple evolutionary strategy (ES)~\cite{hansen2001completely, beyer2002evolution} to approximate the gradient:
\begin{align} \label{eq_gradient_approximation_es}
\overline{\nabla}_{\vx} H(\vx,t) = \frac{1}{\sigma K} \sum_{k=1}^{K} (H(\vx + \sigma \vu^{(k)},t) -  H(\vx,t)) \vu^{(k)},
\end{align}
where $\{\vu^{(1)},\ldots,\vu^{(K)}\}$ are $K$ i.i.d. $d$-dimensional Gaussian vectors sampled from $N(0,I_d)$ and $\sigma$ is a fixed control parameter. This simple gradient estimation is also closely related to Gaussian smoothing~\cite{nesterov2017random,gao2022generalizing}. 

For the Gaussian homotopy function (\ref{gaussian_homotopy}), according to \citep{nesterov2017random}, its gradient can be written as:   
\begin{align} \label{eq_gradient_gh}
&\nabla_{\vx} GH(\vx,t) \\
&= \frac{1}{\beta (1-t)} \bbE_{\vu \sim N(0,I_d)}([f(\vx + \beta (1-t) \vu) - f(\vx)]\vu). \nonumber
\end{align}
Therefore, its gradient can be approximated by:
\begin{align} \label{eq_gradient_approximation_gh}
&\overline{\nabla}_{\vx} GH(\vx,t) \\
&= \frac{1}{\beta (1-t)K} \sum_{k=1}^{K} (f(\vx + \beta (1-t) \vu^{(k)}) - f(\vx))\vu^{(k)}, \nonumber
\end{align}
where $\{\vu^{(1)},\ldots,\vu^{(K)}\} \sim N(0,I_d)$ are $K$ i.i.d. Gaussian vectors as in the ES approximate gradient (\ref{eq_gradient_approximation_es}), while we now only query the value for the original function $f$ but not the homotopy function $H(\vx, t)$. A similar zeroth-order approximation with batch size $K=1$ has been used and analyzed in \citet{iwakiri2022single} for the SLGH algorithm. 

\subsection{Optional Local Search}

\begin{figure}[h]
    \centering
    \includegraphics[width= 1 \linewidth]{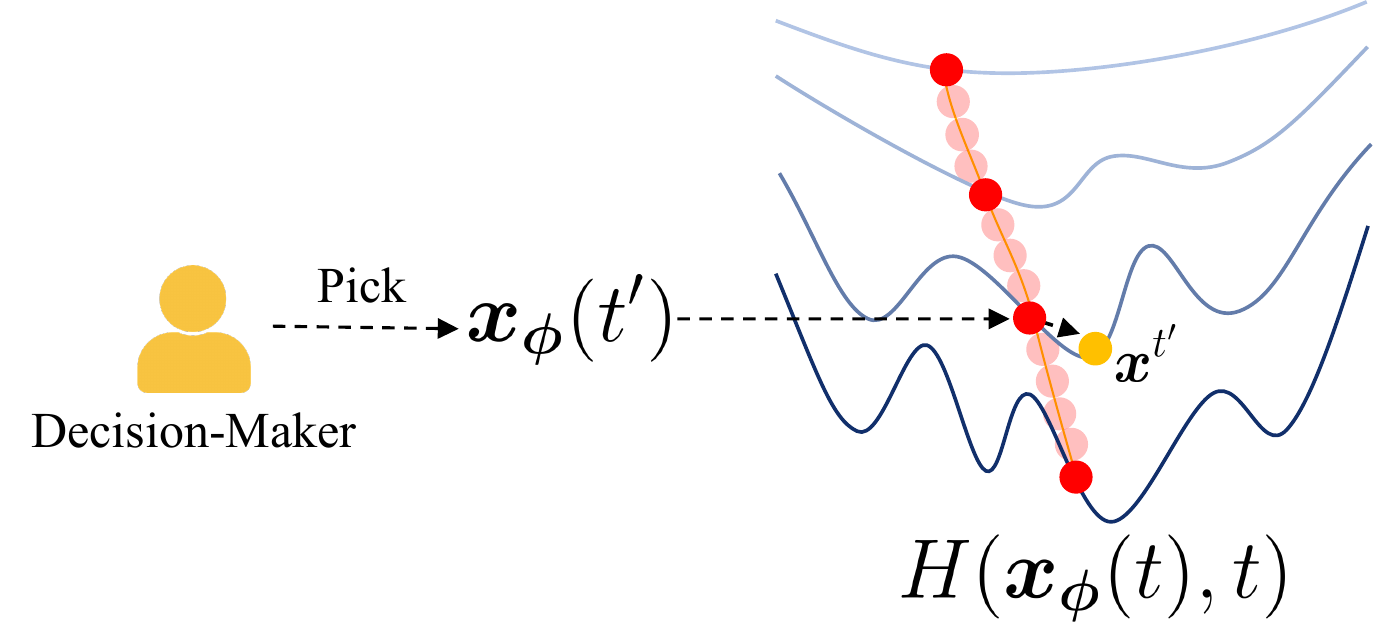}
    \caption{\textbf{Optional Local Search:} Decision-makers can easily pick their favorite solutions from the learned continuation path. An optional local search can help to find a better solution if needed.}
    \label{fig_local_search}
\end{figure}

The previous subsections mainly focus on learning the whole continuation path. In some applications, the decision-maker might only be interested in a single solution, such as $\vx_{\vphi}(t=1)$ for the original optimization problem. By learning the whole continuation path $\vx_{\vphi}(t)$ for all homotopy levels together, CPL actually exchanges the information among different homotopy subproblems simultaneously via the path model. In this case, our learned continuation path model can act as a warm start for any homotopy subproblem, which is more flexible than the gradual unidirectional information passing in classical homotopy optimization. An optional fast local search could help to find a better solution as in \textbf{Algorithm~\ref{alg_path_learning}} and Figure~\ref{fig_local_search}. This step is equal to the final iteration of the classical homotopy optimization algorithm. In the next section, we empirically show that CPL can indeed find better initial solutions. 

\begin{figure*}[t]
\centering
\subfloat[Ackley]{\includegraphics[width = 0.16\textwidth]{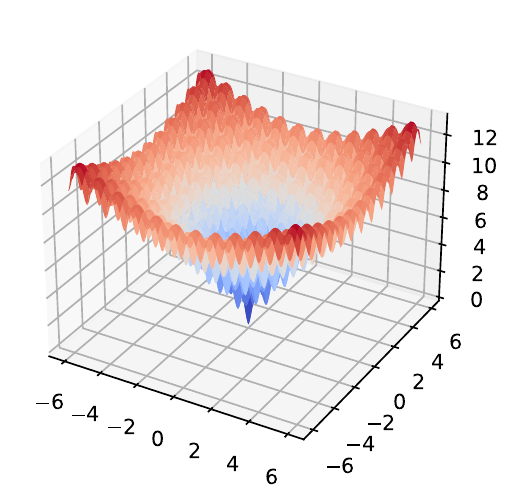}} 
\subfloat[GH Ackley]{\includegraphics[width = 0.16\textwidth]{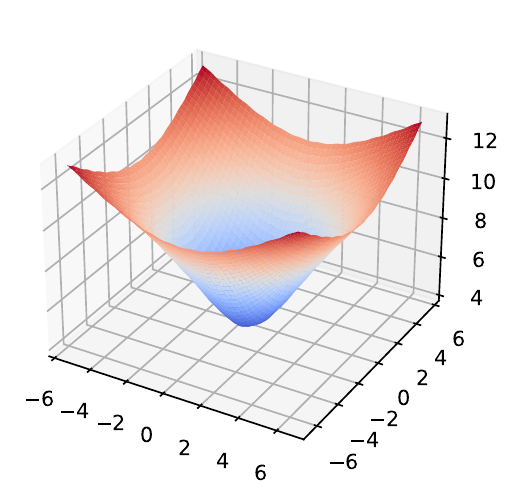}} 
\subfloat[Rosenbrock]{\includegraphics[width = 0.16\textwidth]{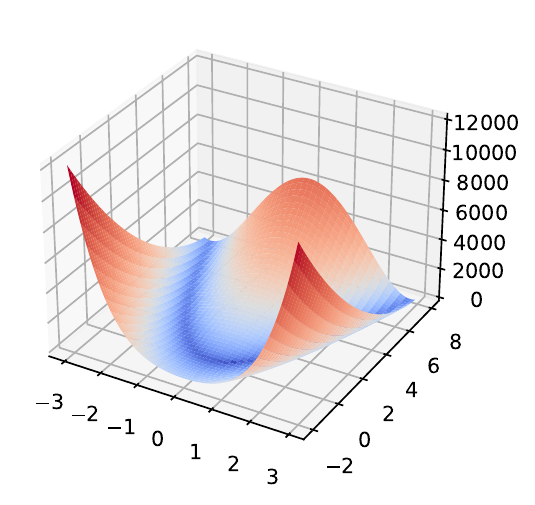}} 
\subfloat[GH Rosenbrock]{\includegraphics[width = 0.16\textwidth]{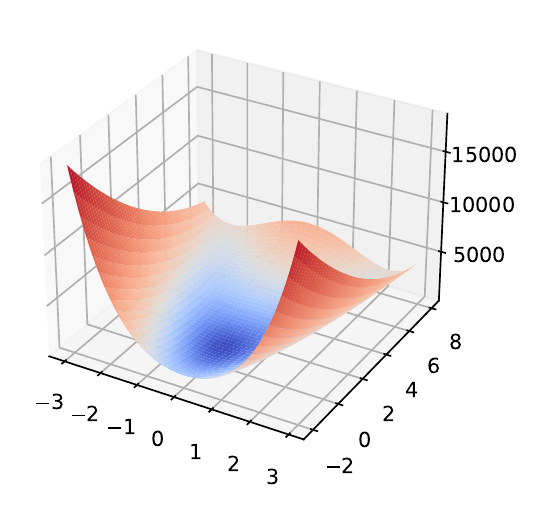}} 
\subfloat[Himmelblau]{\includegraphics[width = 0.16\textwidth]{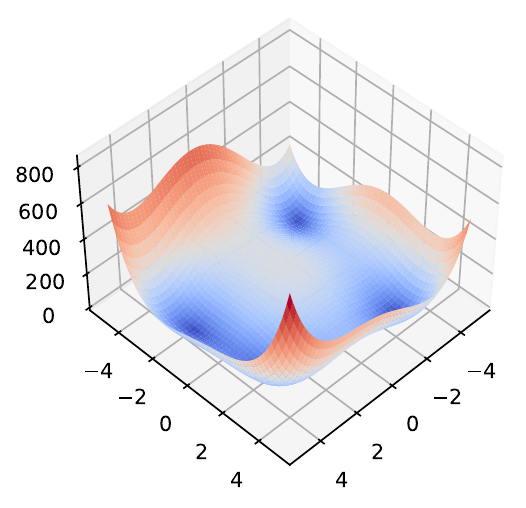}}
\subfloat[GH Himmelblau]{\includegraphics[width = 0.16\textwidth]{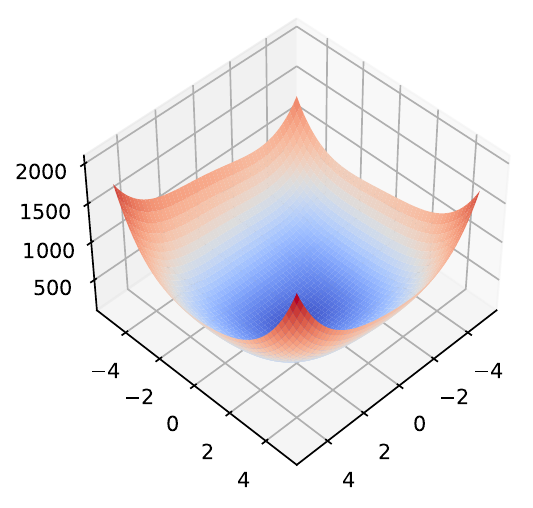}}
\caption{The original and Gaussian homotopy (GH) versions of the Ackley (a,b), Ronsenbrock (c,d), and Himmelblau (e,f) optimization problems. The original optimization problems are all non-convex, and the GH surrogate problems are much smoother and easier to solve.}
\label{fig_synthetic_problems}
\end{figure*}

\begin{table*}[t]
\centering
\caption{Results of gradient descent algorithm, Gaussian homotopy optimization algorithms, and our proposed continuation path learning (CPL) algorithm on three widely-used optimization problems. The optimal value is $0$ for all problems. CPL can obtain the best solutions for all problems with the same number of function evaluations.}
\label{table_synthetic_problem}
\begin{tabular}{llccc}
\toprule
\multicolumn{2}{l}{Algorithms}                                   & Ackley                                 & \multicolumn{1}{l}{Rosenbrock}          & \multicolumn{1}{l}{Himmelblau}                                     \\ \midrule
Gradient Descent           & GD                                  & 12.63                                  & 0.2840                                  & $1.6 \times 10^{-4}$                                               \\ \midrule
Homotopy Optimization      & GradOpt($\gamma = 0.5$)             & 0.014                                  & 0.0336                                  & 14.14                                                              \\
                           & GradOpt($\gamma = 0.8$)             & 0.081                                  & 0.0370                                  & 80.51                                                              \\
                           & $\text{SLGH}_{r}$($\gamma = 0.995$) & 6.650                                  & 0.0327                                  & $6.9 \times 10^{-5}$                                               \\
                           & $\text{SLGH}_{r}$($\gamma = 0.995$) & 0.017                                  & 0.0419                                  & 0.21                                                               \\ \midrule
Continuation Path Learning & CPL($95\%$ path model training)                     & 0.022                                  & 0.0421                                  & $1.7 \times 10^{-3}$                                               \\
                           & CPL(path model + local search)      & \cellcolor[HTML]{67FD9A}\textbf{0.006} & \cellcolor[HTML]{67FD9A}\textbf{0.0018} & \cellcolor[HTML]{67FD9A}\bm{$2.3 \times 10^{-6}$} \\ \bottomrule
\end{tabular}
\end{table*}

\section{Bridging Homotopy Optimization and Parametric Optimization}

Our proposed continuation path learning approach indeed provides a novel view to bridge homotopy optimization and parametric optimization. We discuss several interesting connections in this section.

\subsection{CPL as Parametric Optimization}

A general parametric optimization problem is defined as:
\begin{eqnarray}\label{eq_parametric_opt}
\min_{\vx \in \mathcal{X}} f(\vx, \vbeta),
\end{eqnarray} 
where $\vx \in \mathcal{X}$ is the decision variable, $\vbeta \in \mathcal{B}$ is the problem parameter (also called the context), and $f: \mathcal{X} \times \mathcal{B} \rightarrow \bbR$. Classical works on parametric optimization mainly focus on the sensitivity of objective value to the problem parameter~\citep{bank1983non,bonnans2013perturbation,still2018lectures}. A typical metric is the value function 
\begin{eqnarray}\label{eq_value function}
v(\vbeta) = \min_x f(\vx, \vbeta)
\end{eqnarray} 
that describes the change of optimal value with respect to the problem parameter $\vbeta$.     

In our work, we propose to build a model to approximate the whole continuation path $\vx_{\vphi^*}(t) = \vx^*(t) = \argmin_{\vx} H(\vx, t)$ with every homotopy level $t$. If we treat the homotopy level $t$ as the problem parameter, we can define the value function for homotopy optimization
\begin{eqnarray}
v(t) = \min_{\vx} H(\vx, t) = H(\vx = \vx_{\vphi^*}(t), t)
\label{eq_parametric_optimization}
\end{eqnarray}
for all valid $t$. With CPL's model-based reformulation, we can now use the well-studied parametric optimization approach as a novel view to analyze homotopy optimization methods.

In addition, the classic value function approach mainly addresses the change of optimal value $v(\vbeta)$ with respect to the parameter $\vbeta$. The direct differentiation of $v(\vbeta)$ could be difficult since the optimal solution $\vx$ is often unavailable~\citep{mehmood2020automatic,mehmood2021differentiating}. By modeling $\vx^*(t) = x_{\vphi^*}(t)$, the homotopy objective $v(t) = H(\vx = \vx_{\vphi^*}(t), t)$ can be easily differentiated and optimized by gradient-based optimization methods. The solution model $x_{\vphi}(t)$ (in addition to the objective value) may also provide useful information to support decision-making. It could be interesting to extend the solution model method for a general value function approach.

\subsection{Connection to Amortized Optimization}

There is an exciting and important research direction on learning to optimize (or called Amortized Optimization)~\citep{amos2022tutorial,chen2022learning}, which focuses on making implicit or explicit inferences from a given problem context to its solution. Some recent works directly predict the optimal solution from problem parameters with specific structures~\citep{liu2022teaching}. Conceptually, they are similar to the value function approach but also with a solution model.

A recent work~\citep{li2023homotopy} proposes a classic iterative homotopy optimization method to accelerate the learning-to-optimize approach. Our proposed CPL method can be useful to further improve its performance. If we treat the homotopy level as an additional problem parameter, it is possible to learn the continuation path for a set of problems via a single model. The viewpoint of the value function approach may also provide useful insight for designing better methods.

\subsection{Problem Reformulation and Difficulty}

Our proposed CPL approach reformulates the original optimization problem into continuation model training which might have more parameters to optimize, but we believe it can deal with the homotopy optimization more easily. First of all, there could be infinite homotopy levels and corresponding intermediate solutions for the original problem, which is hard to handle using classic iterative optimization approaches. Our proposed CPL method can learn the whole continuation path via a single model, which is a novel and principled way to deal with this problem. In addition, the existing homotopy optimization methods could be sensitive to the progressive schedule design, and also suffer from high computational overheads due to the iterative optimization structure~\citep{iwakiri2022single}. In contrast, our model-based CPL approach can be easily trained by a gradient-based optimization algorithm and obtain promising results. The reason why a large deep neural network can be easily trained by stochastic gradient descent is still an open research question. For the problem transformation in CPL, some findings from the smooth parametrization work such as \citet{levin2022effect} would be useful for further studies.

\section{Experimental Studies}
\label{sec_experiment}

In this section, we empirically evaluate different aspects of our proposed continuation path learning (CPL) method for solving non-convex optimization, noisy regression, and neural combinatorial optimization problem. Due to the page limit, we only report the main results in this section. Detailed experimental settings, extra experimental results, and more discussions for each problem can be found in Appendix~\ref{supp_sec_synthetic_problem},\ref{supp_sec_nonlinear_regression}, and~\ref{supp_sec_combinatorial_optimization} respectively.

\begin{figure*}[t]
\centering
\subfloat[F1]{\includegraphics[width = 0.25\textwidth]{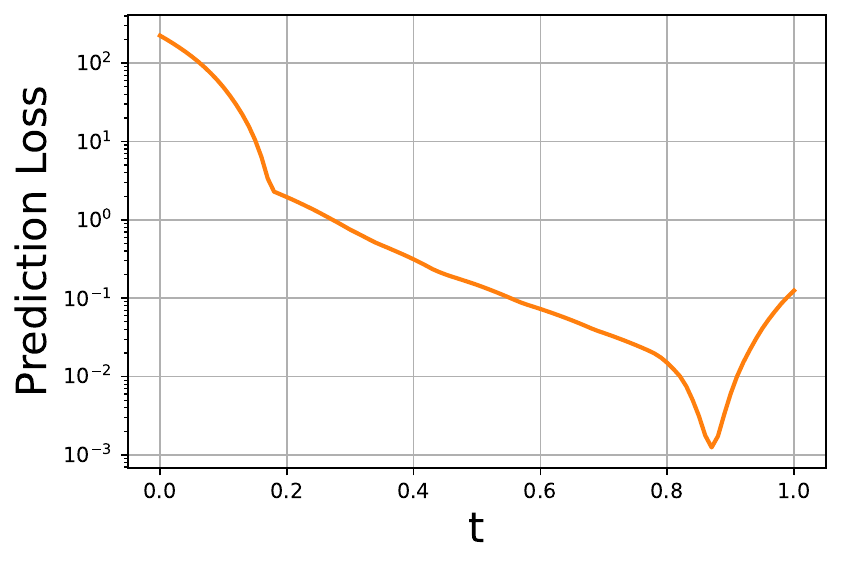}} 
\subfloat[F2]{\includegraphics[width = 0.25\textwidth]{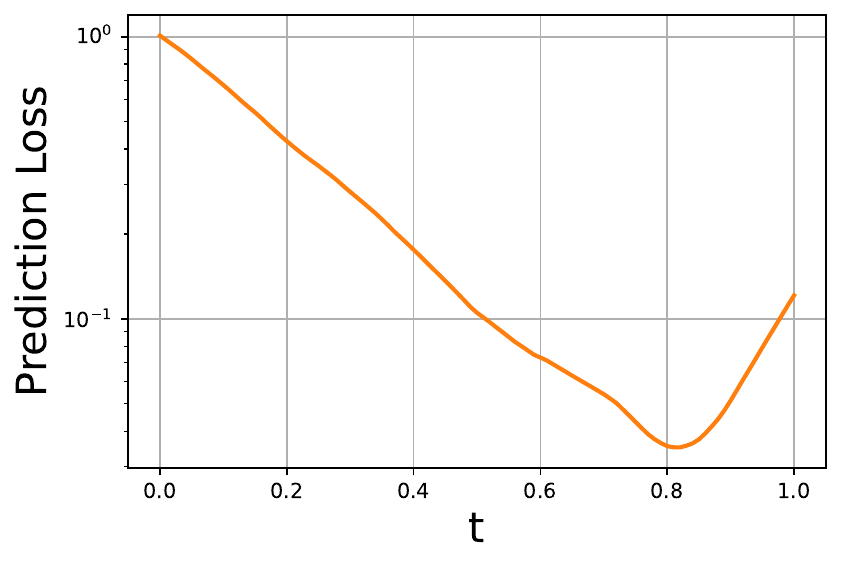}} 
\subfloat[F3]{\includegraphics[width = 0.25\textwidth]{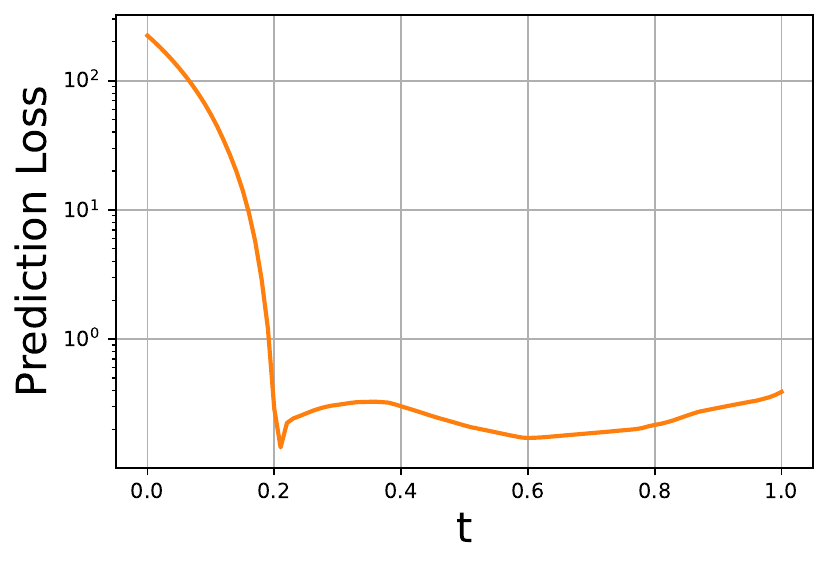}} 
\subfloat[F4]{\includegraphics[width = 0.25\textwidth]{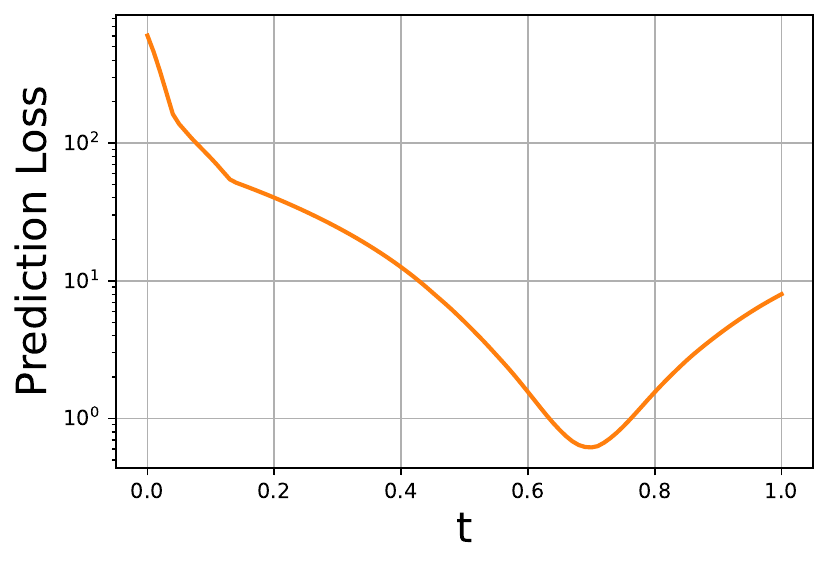}} 
\caption{Prediction performance v.s. homotopy level $t$ on four different noisy nonlinear regression problems. CPL can successfully learn the whole continuation path for all problems. With the learned path, the decision-maker can easily locate the optimal homotopy level $t$ with the minimum prediction loss for each problem.}
\label{fig_regression}
\end{figure*}

\subsection{Non-convex Optimization}

We first test CPL's performance on three widely-used synthetic test benchmark problems, namely the Ackley function~\cite{ackley1987connectionist}, the Rosenbrock function~\cite{rosenbrock1960automatic}, and the Himmelblau function~\cite{himmelblau1972applied}. The original and Gaussian homotopy versions of these functions are shown in Figure~\ref{fig_synthetic_problems}. The original optimization functions are non-convex and hence hard to be directly optimized by simple gradient descent algorithms. In contrast, their Gaussian homotopy versions are much smoother and easier to optimize.

We mainly follow the experimental setting from \citet{iwakiri2022single}, and compare CPL with a simple gradient descent algorithm (GD), a classical Gaussian optimization algorithm (GradOpt)~\cite{hazan2016graduated} with two smoothing parameters ($\gamma = 0.5$ and $0.8$), and the recently proposed single loop Gaussian homotopy algorithm with fixed ratio update ($\text{SLGH}_{r}$)~\cite{iwakiri2022single} with $\gamma = 0.995$ and $0.999$. The total numbers of function evaluations are $1,000$, $20,000$, and $2,000$ for the Ackley, Rosenbrock, and Himmelblau optimization problem respectively. For CPL, since the goal is to optimize the original optimization problem, we use $95\%$ function evaluations for path model training and the rest $5\%$ for gradient-based local search with initial solution $\vx_{\vphi}(t=1)$. In other words, CPL has the same number of function evaluations as other methods.  

According to the results reported in Table~\ref{table_synthetic_problem}, the classic homotopy optimization methods are sensitive to the scheduling design and hyperparameters, which would be hard to tune for a new problem. Indeed, all the homotopy optimization methods are carefully fine-tuned as in \citet{iwakiri2022single}, but no single method can consistently achieve good performance on all problems. In contrast, our proposed CPL method with only the model training step ($95\%$ evaluations) can already generate promising $\vx_{\vphi}(t=1)$ solutions that have similar performances with other homotopy optimization algorithms. With extra gradient-based local search (the rest $5\%$ evaluations), CPL can achieve significantly better solutions for all test problems. These results confirm that learning the whole continuation path in a collaborative manner with knowledge transfer can be helpful for solving the original complicated problem. It is a strong alternative to the classical homotopy optimization methods.

\subsection{Noisy Regression}

In this subsection, we consider the following noisy nonlinear regression problem:
\begin{align} \label{eq_nosiy_regression}
\min_{\valpha \in \bbR^d} t ||\hat \vy - \psi(\hat \vX)\valpha||^2_2 + (1-t)||\valpha||_2,
\end{align}
where $\hat \vX \in \bbR^{n \times p}$ is a matrix of predictors, $\hat \vy \in \bbR^n$ is a noisy response vector, and $\psi: \bbR^p \rightarrow \bbR^d$ is a nonlinear mapping to the feature space. Given a noisy data set with $n$ data points $\{\hat \vX, \hat \vy\}$, our goal is to find the optimal parameters $\valpha^* = \argmin_{\valpha} ||\vy - \psi(\vX)\valpha||^2_2$ for the noiseless $\{\vX, \vy\}$. The noisy regression problem (\ref{eq_nosiy_regression}) is a proper homotopy surrogate controlled by the continuation level $t$. When $t = 0$, the problem reduces to $\min_{\valpha \in \bbR^d} ||\valpha||_2$ which has a trivial solution $\valpha = \boldsymbol{0}^d$. When $t = 1$, it is the standard regression problem $\min_{\valpha \in \bbR^d} ||\hat \vy - \psi(\hat \vX)\valpha||^2_2$ without the regularization term $||\valpha||_2$, which could overfit to the noisy data. To have the best prediction performance on the noiseless $\{\vX, \vy\}$, we need to find the solution for the homotopy optimization problem (\ref{eq_nosiy_regression}) with a proper but unknown $t \in \mathcal{T} = [0,1]$.

Our proposed CPL method can learn the whole continuation path for this problem. We build a simple fully connected neural network $\valpha(t) = h_{\vphi}(t)$ as the path model which maps any valid homotopy level $t$ to its solution $\valpha(t)$, and reformulate the noisy regression problem into:
\begin{align} \label{eq_nosiy_regression_cpl}
\min_{\vphi} t ||\hat \vy - \psi(\hat \vX)h_{\vphi}(t)||^2_2 + (1-t)||h_{\vphi}(t)||_2.
\end{align}
Then the path model can be trained by simple gradient descent to obtain the optimal model parameters $\vphi$. We learn the continuation path for four different noisy regression problems and report their results in Figure~\ref{fig_regression}. CPL successfully learns the continuation path for all problems, which can be used to directly locate the optimal homotopy level $t$. The problem details can be found in Appendix~\ref{supp_sec_nonlinear_regression}.

\begin{figure*}[t]
\centering
\subfloat[t = 0]{\includegraphics[width = 0.18\textwidth]{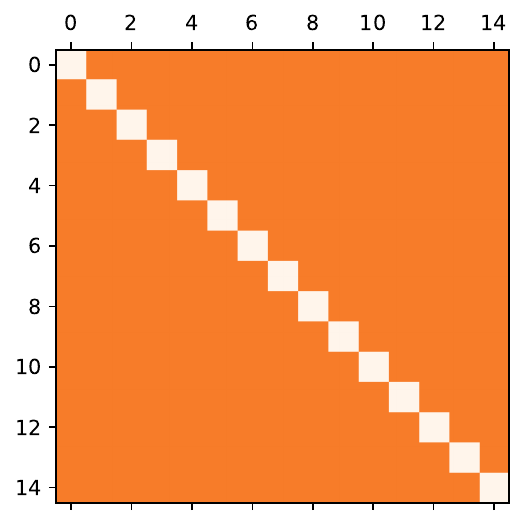}} 
\subfloat[t = 0.25]{\includegraphics[width = 0.18\textwidth]{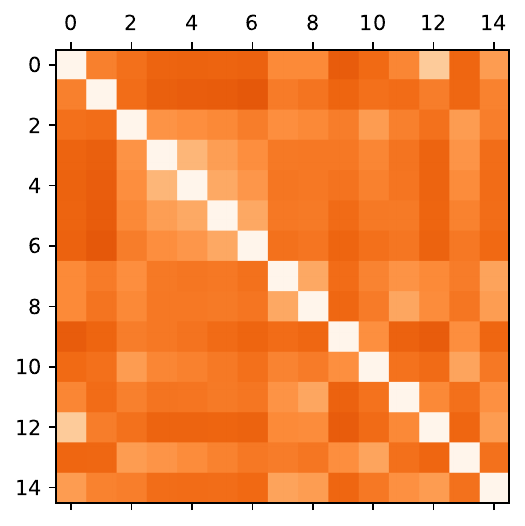}} 
\subfloat[t = 0.5]{\includegraphics[width = 0.18\textwidth]{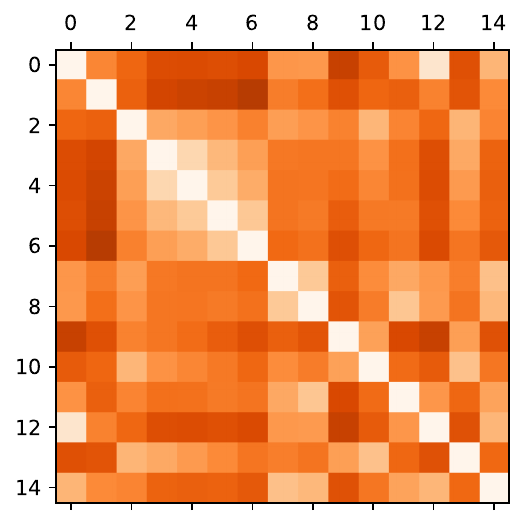}} 
\subfloat[t = 0.75]{\includegraphics[width = 0.18\textwidth]{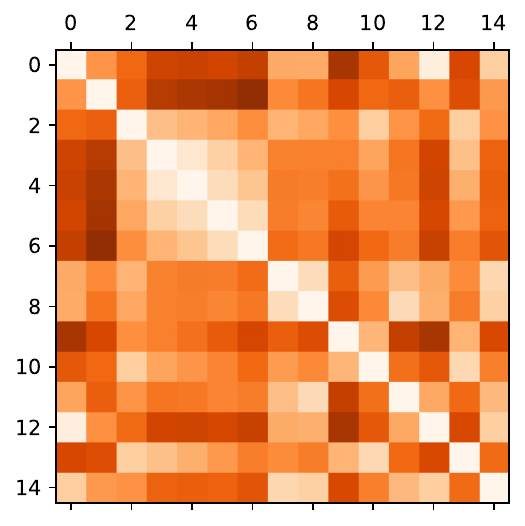}} 
\subfloat[t = 1]{\includegraphics[width = 0.18\textwidth]{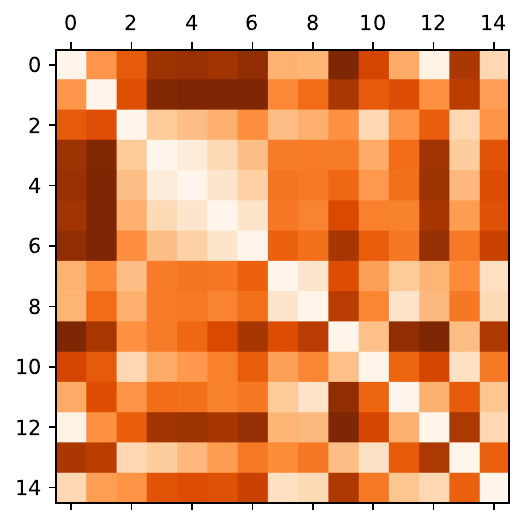}} 
\caption{ \textbf{Homotopy Subproblems for TSP}: \textbf{(a)} When $t =0$, we have the smoothest subproblem where all the distances are the same. In this case, all valid tours will have the same length and be equally optimal, so solving TSP could be trivial. \textbf{(b)-(d)} When $t$ increases, we have more rugged subproblems that gradually transform back to the original problem. \textbf{(e)} When $t = 1$, we have the original problem.}
\label{fig_continuation_tsp}
\vspace{-0.1in}
\end{figure*}

\begin{table*}[t]
\centering
\caption{Results of generalization performance on the realistic TSPLib instances (with $51$ to $200$ cities). All learning-based methods are trained on synthetic and uniformly distributed TSP instances with $100$ cities. The full table can be found in Table~\ref{table_TSPLib}.}
\label{table_tsplib_short}
\begin{tabular}{llllllll}
\toprule
            & OR-Tools & Wu et al.~\cite{wu2021learning} & DACT   & DACT(long)   & AM-S    & POMO   & CPL    \\ \midrule
Optimal Gap & 3.34\%   & 4.17\%    & 3.90\% & 2.07\% & 22.83\% & 2.15\% & 1.72\% \\ \bottomrule
\end{tabular}
\vspace{-0.1in}
\end{table*}

\subsection{Neural Combinatorial Optimization}
\label{sub_sec_nco}

The CPL method can also improve the generalization performance for a neural combinatorial optimization (NCO) solver~\cite{vinyals2015pointer,kool2019attention}, which learns to directly predict the solution for a combinatorial optimization problem. We use the popular traveling salesman problem (TSP) to motivate our approach. A Euclidean TSP instance $s$ is a fully connected graph with $n$ nodes (cities) where each city has its own two-dimensional location $\vy$. The traveling cost between two cities $i$ and $j$ can be defined as the distance $c_{ij} = ||\vy_i - \vy_j||_2$. The goal of TSP is to find a valid tour with the shortest cost to visit all cities exactly once and then return to the starting city. We can represent a valid tour as a permutation of all cities $\vpi = (\pi_1, \cdots, \pi_i, \cdots, \pi_n), \pi_i \in \{1,\cdots, n\}$, and the objective is to find the optimal tour to minimize:
\begin{equation}\label{eq_tsp}
l(\vpi|s) =  c_{\pi_n\pi_1} + \sum_{i = 1}^{n - 1} c_{\pi_i\pi_{i+1}}.
\end{equation}
We can construct a smoother homotopy subproblem by gradually changing the cost between city pairs~\cite{coy2000computational}:
\begin{equation}\label{eq_concave_smoothing}
\hat c_{ij}(t) =c_{ij}^t, \quad t \in \mathcal{T} = [0,1].
\end{equation}
We always normalize the original cost matrix such that $c_{ij} \in [0,1]$ and hence all $\hat c_{ij}(t) \in [0,1]$, and we also normalize the smoothed cost matrix to have the same mean with the original cost matrix $\sum \hat c_{ij}(t) = \sum c_{ij} = \sum \bar c$. The cost matrices with different values of $t$ are shown in Figure~\ref{fig_continuation_tsp}. With the smoothed costs, we can define the continuation function of the tour $\vpi$ for problem instance $s$ as:
\begin{equation}\label{eq_smooth_tsp} 
H(\vpi,t|s) = c^t_{\pi_n\pi_1} + \sum\nolimits_{i = 1}^{n - 1} c^t_{\pi_i\pi_{i+1}}, \quad t \in \mathcal{T} = [0,1].
\end{equation}
For $t = 0$, all valid tours have the same total cost $H(\vpi,t=0|s) = n \bar c$, and the optimization problem becomes trivial to solve. For $t = 1$, we have the original objective function $H(\vpi,t=1|s) = l(\vpi|s)$.

With the above homotopy idea, we can build a CPL-enhanced solver for neural combinatorial optimization. As shown in Figure~\ref{fig_tsp_count}, our model can easily generate multiple solutions on the continuation path for the smoothed subproblems (e.g., $t_m \in [0,1]$) to make a multi-shot prediction that might contain solutions with better generalization performance. By leveraging this property, CPL can obtain a robust generalization performance on unseen TSPlib instances with unseen sizes and distributions as shown in Table~\ref{table_tsplib_short}. Model details and more results can be found in Appendix.~\ref{supp_sec_combinatorial_optimization}.  

\begin{figure}[t]
    \centering
    \includegraphics[width= 0.8 \linewidth]{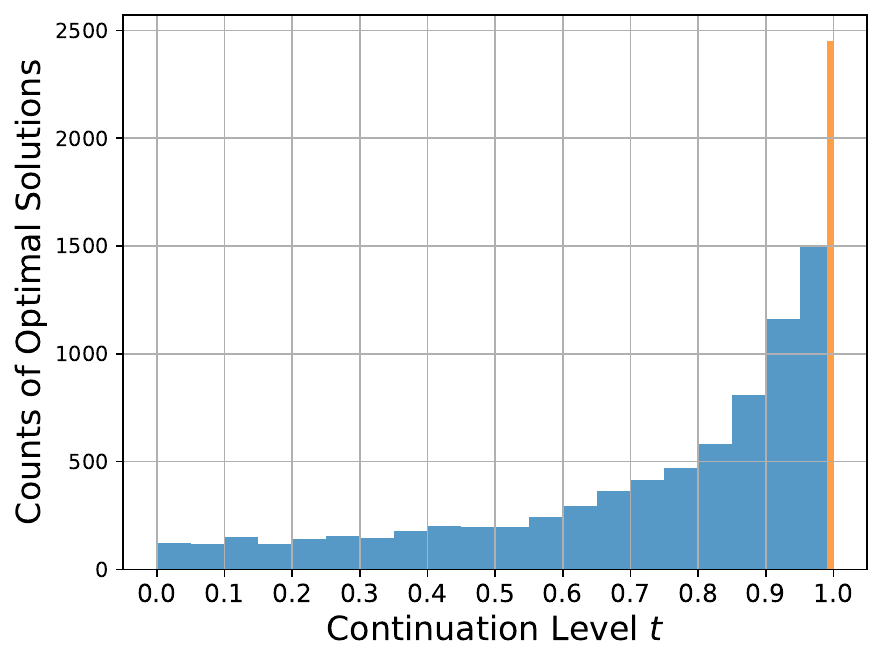}
    \caption{\textbf{CPL inference and counts of optimal solutions.} For $10,000$ random 100-city TSP instances, the model with $t = 1$ can only generate the best solutions for roughly $25\%$ instances. We leverage solutions on the continuation path ($t \in [0,1)$) to achieve better overall performance.}
    \vspace{-0.2in}
    \label{fig_tsp_count}
\end{figure}

\section{Conclusion and Limitation}
\label{sec_conclusion}

\paragraph{Conclusion} We have proposed a novel continuation path learning (CPL) method to approximate the whole continuation path for homotopy optimization. The experimental results have shown that CPL can successfully learn the solution path for different applications. In addition, compared with the classical homotopy optimization method, CPL can achieve similar or even better performance for the original complicated problem. We believe CPL could be a novel and promising method for homotopy optimization.

\paragraph{Limitation}  A limitation of CPL is that we need to build and train a model for learning the continuation path. The suitable model design will mainly depend on the given problem, and some domain knowledge might also be required for efficient model building. Additional theoretical analyses, such as problem transformation and the relation to the value function approach, are important future works.

\section*{Acknowledgements}

This work was supported by the Research Grants Council of the Hong Kong Special Administrative Region, China (Project No. CityU 11215622).

\clearpage
\bibliography{ref_combinatorial_optimization, ref_multiobjective_optimization, ref_multi_task_learning, ref_bayesian_optimization, ref_continuation_and_homotopy, ref_learning_to_optimize}
\bibliographystyle{icml2023}

\newpage
\appendix
\onecolumn


We provide detailed experimental settings, extra experimental results, and more discussions in this Appendix.

\section{Synthetic Test Benchmarks}
\label{supp_sec_synthetic_problem}

\subsection{Problem Definition}

In this experiment, we evaluate the proposed continuation path learning (CPL) method with other homotopy optimization algorithms on the following three widely-used non-convex optimization problems.

\textbf{Ackley Optimization Problem~\cite{ackley1987connectionist}:}
\begin{align} \label{eq_supp_ackley}
f(x,y) = -20 e^{-0.2\sqrt{0.5(x^2 + y^2)}} - e^{0.5(\cos{2\pi x} + \cos{2\pi y})} + e + 20.
\end{align}

\textbf{Rosenbrock Optimization Problem~\cite{rosenbrock1960automatic}:}
\begin{align} \label{eq_supp_rosenbrock}
f(x,y) = 100(y-x^2)^2 + (1-x)^2.
\end{align}

\textbf{Himmelblau Optimization Problem~\cite{himmelblau1972applied}:}
\begin{align} \label{eq_supp_himmelblau}
f(x,y) = (x^2 + y -11)^2 + (x + y^2 - 7)^2.
\end{align}

\subsection{Experimental Setting}

We compare CPL with a simple gradient descent algorithm (GD), a classical Gaussian optimization algorithm (GradOpt)~\cite{hazan2016graduated} with two smoothing parameters ($\gamma = 0.5$ or $0.8$), and the recently proposed single loop Gaussian homotopy algorithm with fixed ratio update ($\text{SLGH}_{r}$)~\cite{iwakiri2022single} with $\gamma = 0.995$ or $0.999$. We report the results of these algorithms with fine-tuned hyperparameters from \citet{iwakiri2022single}.   

For our proposed CPL method, we build a simple fully connected (FC) neural network as the continuation path model. It has two hidden layers each with $128$ hidden nodes. Since the model's gradient can be decomposed with the simple chain rule as in Section~\ref{subsec_path_learning}, it can be easily optimized by the (zeroth-order) gradient descent algorithm similar to other homotopy optimization algorithms. CPL can learn to approximate the whole continuation path simultaneously, hence it does not require any predefined continuation schedule or smoothing parameter. 

We use Gaussian homotopy (GH) as the homotopy method and mainly follow the experimental setting from \citet{iwakiri2022single} for each optimization problem.

\paragraph{Ackley} The Ackley optimization problem does not have an analytical form for its Gaussian homotopy function, and hence we use the zeroth-order method to approximate its Gaussian homotopy gradient (\ref{eq_gradient_approximation_gh}). The total number of function evaluations is $1,000$ for all algorithms. CPL uses $950$ evaluations for path model training and $50$ evaluations for final local search with homotopy level $t=1$.   

\paragraph{Rosenbrock} The Gaussian homotopy function of the Rosenbrock optimization problem has the following analytical form~\cite{mobahi2012gaussian,iwakiri2022single}: 
\begin{align} \label{eq_rosenbrock_gh}
GH(x,y,t) =& \bbE_{u_x,u_y} [f(x + \beta(1-t)u_x, y + \beta(1-t)u_y)] \\
=& 100x^4 + [-200y + 600\beta^2(1-t)^2 + 1]x^2 - 2x + 100y^2 - 200\beta^2(1-t)^2y \nonumber \\ 
&+ 300\beta^4(1-t)^2 + 101\beta^2(1-t)^2 + 1. \nonumber
\end{align}
Therefore, we can use the simple first-order gradient method for CPL and all the other homotopy optimization algorithms. The total number of function evaluations is $20,000$. For CPL, $19,000$ evaluations are used for path model training, and the rest $1,000$ is for the final local search with homotopy level $t=1$. We set $\beta = 1.5$ according to \citet{iwakiri2022single}.  

\paragraph{Himmelblau} Similar to the Ronsebrock problem, since the himmelblau optimization problem is polynomial, it has an analytical Gaussian homotopy function:
\begin{align} \label{eq_himmelblau_gh}
GH(x,y,t) =& \bbE_{u_x,u_y} [f(x + \beta(1-t)u_x, y + \beta(1-t)u_y)] \\
=& x^4 + (2y + 6\beta^2(1-t)^2 - 21)x^2 + (2y^2 + 2\beta^2(1-t)^2 - 14)x  \nonumber \\ 
&+ y^4 + (6\beta^2(1-t)^2 - 13)y^2 + (2\beta^2(1-t)^2 - 22)y \nonumber \\
&+ 6\beta^4(1-t)^4 - 34\beta^2(1-t)^2 + 170 \nonumber
\end{align}
We also use the first-order gradient method to optimize this problem for all algorithms. The total number of function evaluations is $2,000$. CPL has $1,900$ evaluations for model training and $100$ for the final local search with $t = 1$. The parameter $\beta$ is set to $2$ as in \citet{iwakiri2022single}.

\subsection{Computational Cost}

\begin{table}[h]
\centering
\caption{Runtime of gradient, descent, SLGH, and CPL with CPU or GPU.}
\begin{tabular}{lccc}
\toprule
Problem          & \multicolumn{1}{l}{Ackley} & \multicolumn{1}{l}{Himmelblau} & \multicolumn{1}{l}{Rosenbrock} \\
\# Iteration     & 1,000                      & 2,000                          & 20,000                         \\ \midrule
Gradient Descent & 0.9s                       & 1.6s                           & 11.2s                          \\
SLGH             & 0.9s                       & 1.7s                           & 12.6s                          \\ \midrule
CPL (CPU)        & 0.9s                       & 1.8s                           & 13.2s                          \\
CPL (GPU)        & 1.6s                       & 2.8s                           & 15.4s                          \\ \bottomrule
\end{tabular}
\end{table}

The CPL model training can be easily done by highly efficient deep learning frameworks such as PyTorch. For these non-convex optimization benchmarks, depending on the number of iterations, CPL typically needs 1 to 15 seconds to train the model while the run times for other model-free methods are 1 to 13 seconds. It should be pointed out that CPL can learn the whole continuation path while the other methods are to find a single final solution. In addition, the CPL training on GPU (RTX-3080) is actually slower than its counterpart on CPU. For such a small model, the reason could be the cost of data transformation from RAM to GPU is larger than the speedup of GPU over CPU. 

\subsection{Effect of the Model Size}

\begin{table}[h]
\tiny
\centering
\caption{CPL performance on the original problem with different model sizes.}
\begin{tabular}{l|c|ccc|ccc|ccc}
\toprule
           & Baseline & \multicolumn{3}{c|}{Single Hidden Layer} & \multicolumn{3}{c|}{Two Hidden Layers}                                                                & \multicolumn{3}{c}{Three Hidden Layers}                                                             \\
Model Size & SLGH     & 16          & 128         & 1024         & \multicolumn{1}{l}{16-16} & \multicolumn{1}{l}{128-128 (this paper)} & \multicolumn{1}{l|}{1024-1024} & \multicolumn{1}{l}{16-16-16} & \multicolumn{1}{l}{128-128-128} & \multicolumn{1}{l}{1024-1024-1024} \\ \midrule
Ackley     & 6.650    & 2.605       & 0.478       & 0.261        & 0.089                     & 0.006                                    & 0.005                          & 0.024                        & 0.007                           & 0.006                              \\
Himmelblau & 6.9e-5   & 84.25       & 11.92       & 3.8e-05      & 19.46                     & 2.3e-6                                   & 2.8e-6                         & 8.74                         & 2.1e-6                          & 2.6e-6                             \\
Rosenbrock & 0.0327   & 0.0548      & 0.0409      & 0.0282       & 0.0371                    & 0.0018                                   & 0.0023                         & 0.0220                       & 0.0016                          & 0.0019                             \\ \bottomrule
\end{tabular}
\label{table_model_size}
\end{table}

The performance of CPL depends on the neural network architectures. Indeed, different optimization problems and applications could require different CPL models. A general guideline is that the model should be large enough to learn the continuation path for the given problem. Therefore, we should care about the model size and also problem-specific structure. 

We report the results for different models with various sizes in Table~\ref{table_model_size}. Based on the results, a very small model (such as those with single hidden layers) is not able to learn the whole continuation path, and hence has poor performance. On the other hand, once the model has sufficient capacity (such as those three-layer models with more than 128 hidden units), further increasing the model size will not lead to significantly better performance. Therefore, it is important to choose a suitable model size for a given problem.

\clearpage

\section{Noisy Regression}
\label{supp_sec_nonlinear_regression}

In this experiment, we test our proposed CPL method on the following four different noisy regression problems.

\paragraph{F1} This problem has the following ground truth function relation between $x$ and $y$:
\begin{align} \label{eq_regression_f1}
y = 0.5\sin(x) + 0.3\cos(2x) + 2\cos(3x),
\end{align}
where $x \in [-5,5]$. For the noisy regression, we report $\hat y = y + 0.1\varepsilon$ where $\varepsilon \sim N(0,1)$ as the noise response value, and set $\psi(x) = [\sin(x), \cos(2x), \cos(3x)]$. Therefore, the optimal $\valpha = [0.5, 0.3, 2]$.

\paragraph{F2} This problem has the following ground truth function relation between $x$ and $y$:
\begin{align} \label{eq_regression_f2}
y = \cos(x) + 0.2\sin(2x) + 0.5\sin(3x),
\end{align}
where $x \in [-5,5]$. For the noisy regression, we report $\hat y = y + 0.1\varepsilon$ where $\varepsilon \sim N(0,1)$ as the noise response value, and set $\psi(x) = [\cos(x), \sin(2x), \sin(3x)]$. Therefore, the optimal $\valpha = [1, 0.2, 0.5]$.

\paragraph{F3} This problem has the following ground truth function relation between $x$ and $y$:
\begin{align} \label{eq_regression_f3}
y = e^{0.25x} - 0.2\cos(x) + 0.5\sin(4x),
\end{align}
where $x \in [-5,5]$. For the noisy regression, we report $\hat y = y + 0.1\varepsilon$ where $\varepsilon \sim N(0,1)$ as the noise response value, and set $\psi(x) = [e^{0.25x}, \cos(x), \sin(4x)]$. Therefore, the optimal $\valpha = [1, -0.2, 0.5]$.

\paragraph{F4} This problem has the following ground truth function relation between $x$ and $y$:
\begin{align} \label{eq_regression_f4}
y = 2 * \ln{0.25|x|} + 3\sin(6x) + 4\cos(0.5x),
\end{align}
where $x \in [-5,5]$. For the noisy regression, we report $\hat y = y + 0.1\varepsilon$ where $\varepsilon \sim N(0,1)$ as the noise response value, and set $\psi(x) = [\ln{0.25|x|}, \sin(6x), \cos(0.5x)]$. Therefore, the optimal $\valpha = [2, 3, 4]$.

\clearpage

\section{Continuation Path Learning for Neural Combinatorial Optimization}
\label{supp_sec_combinatorial_optimization}

\subsection{Continuation-based constructive NCO model}
\label{subsec_model}

\begin{figure}[t]
    \centering
    \includegraphics[width= 1.0 \linewidth]{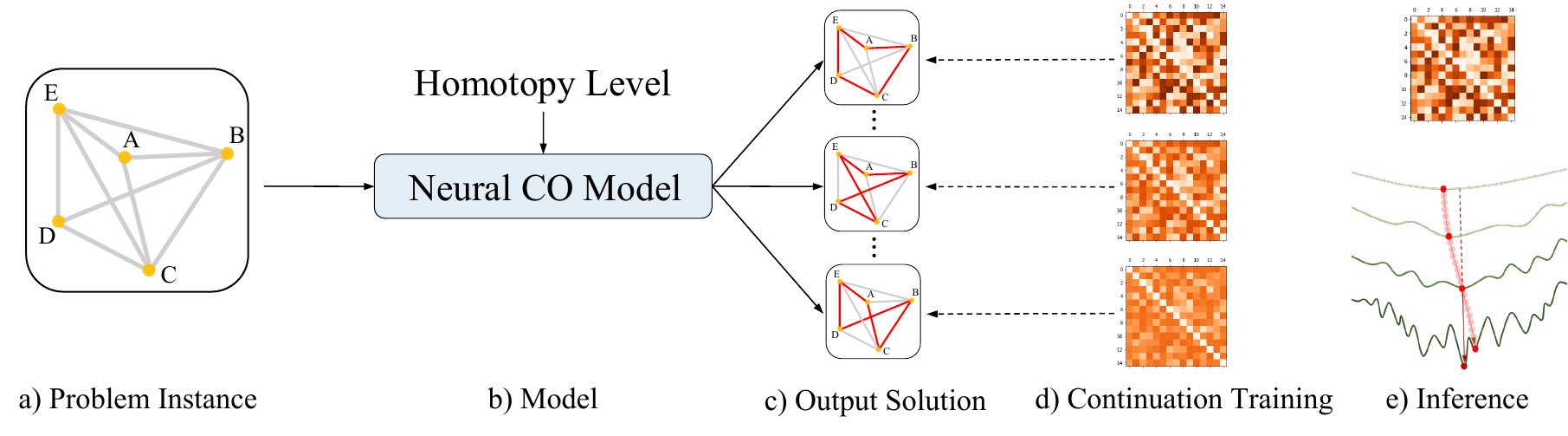}
    \caption{\textbf{Continuation Path Learning Model for Neural Combinatorial Optimization}: Our proposed CPL model can learn to construct different solutions for the smoothed subproblems on the continuation path for a given problem instance. In inference, it leverages these solutions to make a multi-shot prediction for better performance.}
    \label{fig_model_structure}
\end{figure}

In this experiment, we focus on learning the continuation path for constructive neural combinatorial optimization (NCO)~\cite{vinyals2015pointer,bello2016neural}. This approach learns the policy model with parameter $\vtheta$ to construct a solution (e.g., a tour $\vpi = (\pi_1, \cdots, \pi_i, \cdots, \pi_n), \pi_i \in \{1,\cdots, n\}$) for a combinatorial optimization problem instance $s$ (e.g., TSP) in an auto-regressive manner:
\begin{eqnarray}\label{policy_model}
p_{\vtheta}(\vpi | s) = \textstyle\prod_{i = 1}^{n} p_{\vtheta}(\pi_i | s, \pi_{1: i-1}).
\end{eqnarray} 
In contrast to a policy with constant parameters $\vtheta$, we propose to build a policy model with dynamic parameters $\vtheta(t)$ conditioned on the continuation level $t$:
\begin{eqnarray}\label{policy_model_continuation}
p_{\vtheta(t)}(\vpi | s) = \textstyle\prod_{i = 1}^{n} p_{\vtheta(t)}(\pi_i | s, \pi_{1: i-1}),
\end{eqnarray} 
such that it can generate different tours for different smoothed subproblems as in Figure~\ref{fig_model_structure}. By assigning different $t$, we can easily obtain solutions of different smoothed subproblems on the continuation path for a given instance.

The proposed continuation path learning idea and the policy model framework are general and can be used for any constructive NCO model. We use the seminal Attention Model~\cite{kool2019attention} as our constructive model. The recent work shows that only changing (part of) the AM decoder parameters is sufficient to construct solutions with significantly better performance~\cite{hottung2021efficient} or with very different trade-offs among different objectives~\cite{lin2022pareto_combinatorial}. Therefore, we also propose to let only part of the AM decoder parameters depend on the continuation level $t$:
\begin{eqnarray}\label{am_conditioned_decoder}
[W^Q(t), W^{\text{Proj}}(t)] = \textbf{MLP}(t),
\end{eqnarray} 
where $W^Q(t)$ and $W^{\text{proj}}(t)$  are the query and projection parameters for the Multi-Head Attention (MHA)~\cite{vaswani2017attention} layer in the AM decoder. More advanced model structures such as multiplicative interactions~\citep{jayakumar2020multiplicative} and hypernetwork~\citep{schmidhuber1992learning,ha2017hypernetworks} can be used to learn the conditioned parameters, but we find a simple MLP model is good enough for our model. 

To construct a valid tour, our proposed model first tasks the problem instance $s$ (e.g., the graph with $n$ fully connected nodes for TSP) as input to the AM encoder and obtains the $n$ $d$-dimensional node embedding $[\vh_1, \cdots, \vh_n ]$ for each city. For selecting the $i$-th city into the tour, the AM decoder combines the embedding of the first selected node $\vh_{\pi_1}$ and the most current selected node $\vh_{\pi_{i-1}}$ to obtain the query embedding $\vh_Q(t) = W^Q(t)[\vh_{\pi_1}, \vh_{\pi_{i-1}}]$. Following the setting of POMO~\cite{kwon2020pomo}, we do not include an extra graph embedding. The query embedding will be further updated by the multi-head attention with all node embedding:
\begin{eqnarray}\label{multi_head_embedding}
\hat \vh_{Q} = \textbf{MHA}(Q = \vh^{Q}(t), K = W^K[\vh_1,\cdots,\vh_n], V = W^V[\vh_1,\cdots,\vh_n])W^{\text{proj}}(t), 
\end{eqnarray}
where $W^{\text{proj}}(t)$ projects the multi-head output of MHA into an $d$-dimensional embedding. With the query embedding $\hat \vh_{Q}$ and the embedding $\vh_i$ for each city, we can calculate the logit for each city:
\begin{eqnarray}\label{single_head_logit}
\text{logit}_j =  \left\{
             \begin{array}{ll}
             C \cdot \tanh(\frac{\hat \vh_{Q}^T\vh_j}{\sqrt{d}})  &\text{if } j \neq \vpi_{p^{\prime}} \quad \forall p^{\prime} < i,  \\
             -\infty  & \text{otherwise}. 
             \end{array}
\right.
\end{eqnarray}
The logits are further clipped into $[-C, C]$ for all non-selected nodes with $C = 10$ as in \cite{kool2019attention} and all already selected nodes are masked in $-\infty$. The policy model can autoregressively construct a valid tour by following the probability $p_{\vtheta(t)}(\vpi_t = j | s, \vpi_{1: i-1}) = {e^{\text{logit}_j}} / {\sum_{k \in \{ j \neq \vpi_{p^{\prime}} \forall p^{\prime} < i\}} e^{\text{logit}_k}}$.

In the proposed model, the node embedding $[\vh_1,\cdots,\vh_n]$ and the key $K$ and value $V$ of MHA are shared by all continuation level $t \in \mathcal{T} = [0,1]$. Therefore, we only need to calculate them once and then can repeatedly reuse them to construct solutions for different continuation levels $t$. The whole model structure is similar to multi-objective optimization model proposed in \cite{lin2022pareto_combinatorial} but is to learn the continuation path for a single objective function.

\subsection{Model training}
\label{subsec_training}

We have proposed the smoothed subproblems (e.g., (\ref{eq_smooth_tsp})) and the policy model to construct feasible solutions (e.g.,(\ref{policy_model_continuation})) for different continuation levels. Now the goal is to properly train the model so that the generated solutions are on the continuation path (e.g., the optimal solution for each subproblem) of the original problem (\ref{eq_tsp}). The training goal can be defined as:
\begin{eqnarray}\label{expected_cost}
\min_{\vtheta} \mathcal{J}(\vtheta) = \bbE_{\vpi \sim  p_{\vtheta(t)}(\cdot | s),t \sim \mathcal{T}, s \sim \mathcal{S}} H(\vpi,t|s),
\end{eqnarray}
where $\mathcal{S}$ is the set of problem instances, $\mathcal{T} = [0,1]$ is the valid continuation level, and $\vpi \sim  p_{\vtheta(t)}(\cdot | s)$ is the tour generated by the stochastic policy model with respect to the sampled $s$ and $t$. 

We follow our proposed gradient-based continuation path learning method in \textbf{Algorithm~\ref{alg_path_learning}} to train the policy model. The gradient is approximated with REINFORCE~\cite{williams1992simple} and POMO~\cite{kwon2020pomo} rollout:
\begin{eqnarray}\label{reinforce_approx_grad}
\nabla \mathcal{J}(\vtheta) \approx \frac{1}{MBN} \sum_{m=1}^{M} \sum_{i=1}^{B} \sum_{j=1}^{N} [ (H(\vpi_{i}^{j},t_m|s_i) - b(s_i|t_m)) \nabla_{\vtheta(t_m)} \log p_{\vtheta(t_m)}(\vpi_{i}^{j}| s_i)],
\end{eqnarray}
with $M$ sampled continuation levels, $B$ problem instances, and $N$ solutions with diverse started nodes for each instance. We use the shared baseline $b(s_i|t_m) = \frac{1}{N} \sum_{j=1}^{N} H(\vpi_{i}^{j},t_m|s_i)$ for each instance as in POMO~\cite{kwon2020pomo}.

\subsection{Inference from the continuation path}

\begin{figure}[h]
\vspace{-0.2in}
\noindent
\begin{minipage}[t]{0.6\textwidth}
  \centering
  \vspace{0pt}  
\begin{algorithm}[H]
\caption{CPL Inference}
\label{alg_inference}
    \begin{algorithmic}[1]
    \STATE \textbf{Input:} instance $s$, continuation path model $p_{\vtheta(t)}(\vpi | s)$, number of sampled solutions $M$
    \STATE ${t_1,t_2 \ldots, t_M} \sim \mathcal{T} = [0,1]$
    \STATE $\vpi_{t_m} \leftarrow \textbf{GreedyRollout}(p_{\vtheta(t_m)}(\cdot | s)) \quad \forall t_m $ 
    \STATE $\vpi_{\text{best}} = \argmin_{\vpi \in \{\vpi_{t_1},\ldots,\vpi_{t_M}\}} H(\vpi,t = 1|s)$
    \STATE{Output:} $\vpi_{\text{best}}$
    \end{algorithmic}
\end{algorithm}
\end{minipage}%
\hfill
\begin{minipage}[t]{0.35\textwidth}
  \vspace{0pt}
    \begin{figure}[H]
        \centering
        \includegraphics[width= 0.9 \linewidth]{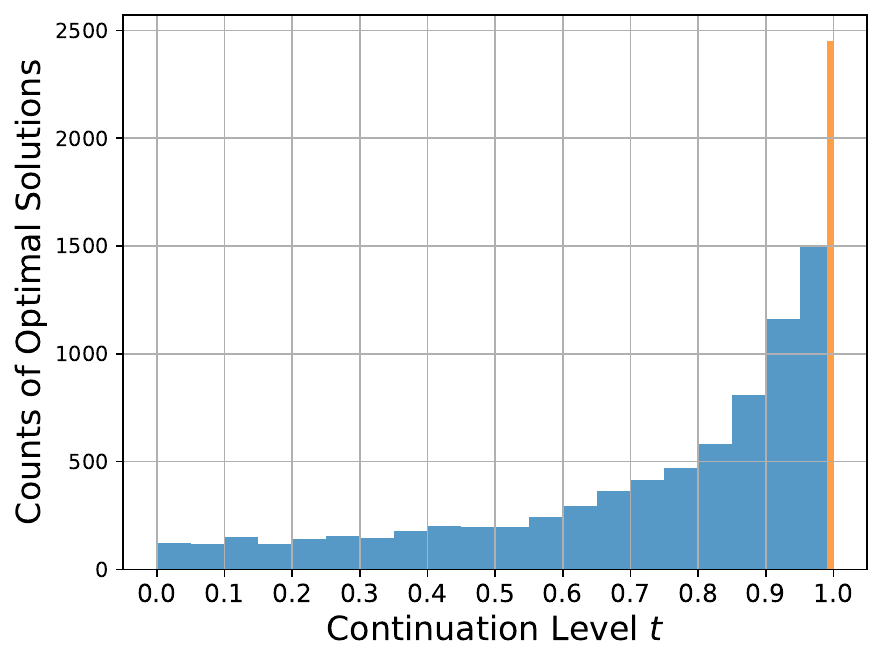}
    \end{figure}
\end{minipage}
\caption{\textbf{CPL inference and counts of optimal solutions.} For $10,000$ random 100-city TSP instances, the model with $t = 1$ can only generate the best solutions for roughly $25\%$ instances. We leverage solutions on the continuation path ($t \in [0,1)$) to achieve better overall performance.}
\label{fig_best_prefs_count}
\end{figure}

Due to the one-shot prediction mechanism and generalization gap, for a new encountered problem instance $s$, the generated solution $\vpi_1$ with $t=1$ might not be the optimal solution for the original objective $H(\vpi,t = 1|s)$ as shown in Figure~\ref{fig_best_prefs_count}. With our proposed model, we can easily generate multiple solutions on the continuation path for the smoothed subproblems (e.g., $t_m \in \mathcal{T} = [0,1]$) to make a multi-shot prediction that might contain solutions with better generalization performance. 

The inference method is summarized in \textbf{Algorithm~\ref{alg_inference}}. For each problem instance $s$, we sample a set of $M$ continuation levels $t_m$, and construct their solutions with POMO rollout~\cite{kwon2020pomo}. The solution with the best original objective value (not the smoothed value) will be chosen as the final solution. 

\subsection{Experimental Setting}

\paragraph{Problem Setting} We evaluate the proposed continuation path learning (CPL) method on both randomly generated TSP and CVRP instances that are widely used in neural combinatorial optimization~\cite{kool2019attention}, as well as more realistic benchmark problems following the suggestion in \cite{accorsi2022guidelines}.

\paragraph{Model setting} We use the Attention Model (AM)~\cite{kool2019attention} as our policy network for construction-based neural combinatorial optimization. The model has a computation-heavy encoder to produce embedding for each node (e.g., city for TSP), and a lightweight decoder to autoregressively generate a valid solution (e.g., tour) based on the node embeddings. The same model structure is used for solving different problems.

We follow the hyperparameters setting from POMO~\cite{kwon2020pomo}. For the encoder, there are $6$ multi-head attention layers with $128$-dimensional embedding, and each has $8$ heads with $16$-dimensional key/value/query embeddings. The fully connected linear sub-layer has dimension $512$ in each attention layer. We build our CPL model based on the new POMO~\cite{kwon2020pomo} codebase~\footnote{https://github.com/yd-kwon/POMO/tree/master/NEW\_py\_ver} under the MIT License, where it uses InstanceNorm for the normalization layers and does not include graph embedding as an output for the decoder. 

Only the parameters for the query token and projection operator are conditioned on the continuation level, and the other parameters are all fixed and shared for all smooth subproblems. In this way, all the key/value embeddings for each node need to be calculated only once, and can be repeatedly reused to construct solutions for different smoothed subproblems.

\begin{figure}[h]
    \centering
    \includegraphics[width= 0.8 \linewidth]{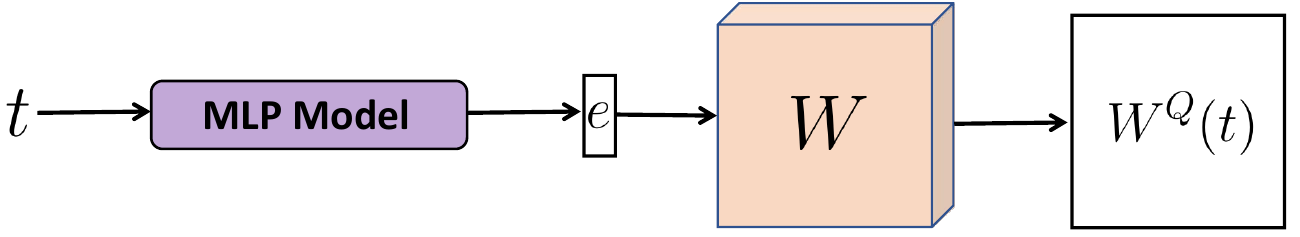}
    \caption{Conditional parameters generation.}
    \label{fig_model_hypernetwork}
\end{figure}

We use a hypernetwork~\cite{ha2017hypernetworks} to generate the continuation conditional parameters $W^Q(t)$ and $W^{\text{proj}}(t)$ as illustrated in Figure~\ref{fig_model_hypernetwork}. It first takes the continuation level $t$ as input for a MLP model to generate a low dimensional embedding $e \in \bbR^d$. Then the embedding $e$ is linearly projected with a parameter tensor $W \in \bbR^{n1 \times n2 \times d}$ to the target parameters $W^Q(t) \in \bbR^{n1 \times n2}$. The number of trainable parameters (e.g., MLP parameters and $W$) is much smaller than a full hypernetwork. In this work, we use a simple $2$-layer MLP with $128$ neurons at each hidden layer, and let $d = 4$ for the embedding.

\paragraph{Training and Inference.} The optimizer we use is Adam with learning rate $\eta = 10^{-4}$, weight decay $\omega = 10^{-6}$ and batch size $B = 64$. At each training epoch, we randomly generate $100,000$ problem instances on the fly as training data, and train the model for $1,000$ epoch. At each batch (e.g., with batch size $B = 64$), we randomly sample $M = 2$ continuation levels, and draw $N$ (e.g., the number of nodes) trajectories for each problem instance with POMO rollout. In this way, the total update step is matched with $2,000$ in POMO without continuation path learning. We train our model on a single RTX $3080$ GPU, which takes roughly $16$ minutes for a training epoch with TSP100. For inference with $M$ continuation levels, we always let $t_1 = 1$ and sample the rest $M-1$ levels from $[0,1)$.

\clearpage

\subsection{Travelling salesman problem (TSP)}

\paragraph{Problem setup} A TSP instance contains the locations of $N$ cities. We need to find a tour with the shortest length to visit all cities once and return to the starting city. 

\paragraph{Instance generation} We follow the setting in AM~\cite{kool2019attention} to randomly generate cities from $[0,1]^2$ with uniform distribution.

\subsection{Capacitated vehicle routing problem (CVRP)}

\paragraph{Problem setup} For a CVRP instance, in addition to the locations of $N$ nodes, each node (city/customer) has a demand $d_i$ to be served. A vehicle with a fixed capacity $C$ from an extra depot node will make multiple round trips to handle the demand for each customer. No demand split is allowed and each city can be visited only once. For a nontrivial and solvable problem instance, we have $\max d_i \leq C < \sum_{i=1}^{N} d_i$. Therefore, the vehicle needs to go back to the depot node to refill its capacity multiple times. The optimal solution is a tour with the shortest length that satisfies the demands of all cities.  
 
\paragraph{Instance generation} Similar to the TSP instance, we uniformly generate the customer nodes and depot nodes from the unit square $[0,1]^2$. For the demands, following AM~\cite{kool2019attention}, we first randomly sample $\hat d_i$ from the discrete set $\{1,2,\ldots,9\}$ and then normalize it to $d_i = \frac{\hat d_i}{D}$, where $D = 30,40,50$ for $N = 20, 50, 100$. The vehicle has a capacity 1. 

A series of smoothed subproblems can be constructed similarly to the TSP instances in Section~\ref{sub_sec_nco}. Now we have an $(N+1) \times (N+1)$ matrix that contains the distance among nodes ($N$ customer nodes and $1$ depot node). The capacity and all demands are unchanged in the smoothed subproblems.

\subsection{Resutls on Random TSP and CVRP Instances}

\begin{table}[ht]
\centering
\caption{Experimental results on TSP and CVRP with random instances}
\begin{tabular}{lccccccccc}
\toprule
\multicolumn{10}{c}{TSP}                                                                                                                                                                            \\ \midrule
\multicolumn{1}{l|}{}                 & \multicolumn{3}{c|}{TSP20}                                      & \multicolumn{3}{c|}{TSP50}                   & \multicolumn{3}{c}{TSP100}                 \\
\multicolumn{1}{l|}{Method}           & Cost                     & Gap     & \multicolumn{1}{c|}{Time}  & Cost  & Gap    & \multicolumn{1}{c|}{Time}   & Cost  & Gap    & Time                      \\ \midrule
\multicolumn{1}{l|}{Concorde}         & 3.83                     & -       & \multicolumn{1}{c|}{(5m)}  & 5.69  & -      & \multicolumn{1}{c|}{(13m)}  & 7.76  & -      & (1h)                      \\
\multicolumn{1}{l|}{LKH3}             & 3.83                     & 0.00\%  & \multicolumn{1}{l|}{(42s)} & 5.69  & 0.00\% & \multicolumn{1}{l|}{(6m)}   & 7.76  & 0.00\% & \multicolumn{1}{l}{(25m)} \\
\multicolumn{1}{l|}{OR Tools}         & \multicolumn{1}{l}{3.86} & 0.94\%  & \multicolumn{1}{l|}{(1m)}  & 5.85  & 2.87\% & \multicolumn{1}{l|}{(5m)}   & 8.06  & 3.86\% & \multicolumn{1}{l}{(23m)} \\ \midrule
\multicolumn{1}{l|}{Neural 2-Opt}     & 3.83                     & 0.00\%  & \multicolumn{1}{c|}{(15m)} & 5.79  & 0.12\% & \multicolumn{1}{c|}{(29m)}  & 7.83  & 0.87\% & (41m)                     \\
\multicolumn{1}{l|}{Wu et al.(T=5k)}  & 3.83                     & 0.00\%  & \multicolumn{1}{c|}{(1h)}  & 5.70  & 0.20\% & \multicolumn{1}{c|}{(1.5h)} & 7.97  & 1.42\% & (2h)                      \\
\multicolumn{1}{l|}{DACT (T=10k)}     & 3.83                     & 0.00\%  & \multicolumn{1}{c|}{(10m)} & 5.70  & 0.00\% & \multicolumn{1}{c|}{(1h)}   & 7.77  & 0.09\% & (2.5h)                    \\ \midrule
\multicolumn{1}{l|}{GCN-beam search}  & 3.83                     & 0.01\%  & \multicolumn{1}{c|}{(12m)} & 5.69  & 0.01\% & \multicolumn{1}{c|}{(18m)}  & 7.87  & 1.39\% & (40m)                     \\
\multicolumn{1}{l|}{AM-greedy}        & 3.84                     & 0.19\%  & \multicolumn{1}{c|}{(1s)}  & 5.76  & 1.21\% & \multicolumn{1}{c|}{(1s)}   & 8.03  & 3.51\% & (2s)                      \\
\multicolumn{1}{l|}{AM-sampling}      & 3.83                     & 0.07\%  & \multicolumn{1}{c|}{(1m)}  & 5.71  & 0.39\% & \multicolumn{1}{c|}{(5m)}   & 7.92  & 1.98\% & (22m)                     \\
\multicolumn{1}{l|}{MDAM-beam search} & 3.84                     & 0.00\%  & \multicolumn{1}{c|}{(3m)}  & 5.70  & 0.03\% & \multicolumn{1}{c|}{(14m)}  & 7.79  & 0.38\% & (44m)                     \\
\multicolumn{1}{l|}{POMO}             & 3.83                     & 0.00\%  & \multicolumn{1}{c|}{(3s)}  & 5.69  & 0.03\% & \multicolumn{1}{c|}{(16s)}  & 7.78  & 0.15\% & (1m)                      \\
\multicolumn{1}{l|}{CPL (4 sols.)}    & \multicolumn{1}{l}{3.83} & 0.00\%  & \multicolumn{1}{c|}{(10s)} & 5.69  & 0.01\% & \multicolumn{1}{c|}{(1m)}   & 7.77  & 0.09\% & (3m)                      \\
\multicolumn{1}{l|}{CPL (8 sols.)}    & 3.83                     & 0.00\%  & \multicolumn{1}{c|}{(19s)} & 5.69  & 0.00\% & \multicolumn{1}{c|}{(2m)}   & 7.77  & 0.08\% & (7m)                      \\ \midrule
\multicolumn{10}{c}{CVRP}                                                                                                                                                                           \\ \midrule
\multicolumn{1}{l|}{}                 & \multicolumn{3}{c|}{CVRP20}                                     & \multicolumn{3}{c|}{CVRP50}                  & \multicolumn{3}{c}{CVRP100}                \\
\multicolumn{1}{l|}{Method}           & Cost                     & Gap     & \multicolumn{1}{c|}{Time}  & Cost  & Gap    & \multicolumn{1}{c|}{Time}   & Cost  & Gap    & Time                      \\ \midrule
\multicolumn{1}{l|}{LKH3}             & 6.12                     & -       & \multicolumn{1}{c|}{(2h)}  & 10.38 & -      & \multicolumn{1}{c|}{(7h)}   & 15.68 & -      & (12h)                     \\
\multicolumn{1}{l|}{OR Tools}         & 6.42                     & 4.84\%  & \multicolumn{1}{c|}{(2m)}  & 11.22 & 8.12\% & \multicolumn{1}{c|}{(12m)}  & 17.14 & 9.34\% & (1h)                      \\ \midrule
\multicolumn{1}{l|}{NeuRewriter}      & 6.16                     & -       & \multicolumn{1}{c|}{(22m)} & 10.51 & -      & \multicolumn{1}{c|}{(18m)}  & 16.10 & -      & (1h)                      \\
\multicolumn{1}{l|}{NLNS}             & 6.19                     & -       & \multicolumn{1}{c|}{(7m)}  & 10.54 & -      & \multicolumn{1}{c|}{(24m)}  & 15.99 & -      & (1h)                      \\
\multicolumn{1}{l|}{Wu et al.(T=5k)}  & 6.12                     & 0.39\%  & \multicolumn{1}{c|}{(2h)}  & 10.45 & 0.70\% & \multicolumn{1}{c|}{(4h)}   & 16.03 & 2.47\% & (5h)                      \\
\multicolumn{1}{l|}{DACT (T=10k)}     & 6.13                     & -0.08\% & \multicolumn{1}{c|}{(35m)} & 10.39 & 0.14\% & \multicolumn{1}{c|}{(1.5h)} & 15.71 & 0.19\% & (4.5h)                    \\ \midrule
\multicolumn{1}{l|}{AM-greedy}        & 6.40                     & 4.45\%  & \multicolumn{1}{c|}{(1s)}  & 10.93 & 5.34\% & \multicolumn{1}{c|}{(1s)}   & 16.73 & 6.72\% & (3s)                      \\
\multicolumn{1}{l|}{AM-sampling}      & 6.24                     & 1.97\%  & \multicolumn{1}{c|}{(3m)}  & 10.59 & 2.11\% & \multicolumn{1}{c|}{(7m)}   & 16.16 & 3.09\% & (30m)                     \\
\multicolumn{1}{l|}{MDAM-beam search} & 6.14                     & 0.18\%  & \multicolumn{1}{c|}{(5m)}  & 10.48 & 0.98\% & \multicolumn{1}{c|}{(15m)}  & 15.99 & 2.23\% & (1h)                      \\
\multicolumn{1}{l|}{POMO}             & 6.14                     & 0.21\%  & \multicolumn{1}{c|}{(5s)}  & 10.42 & 0.45\% & \multicolumn{1}{c|}{(26s)}  & 15.73 & 0.32\% & (2m)                      \\
\multicolumn{1}{l|}{CPL (4 sols.)}    & 6.13                     & 0.14\%  & \multicolumn{1}{c|}{(16s)} & 10.40 & 0.26\% & \multicolumn{1}{c|}{(1m)}   & 15.72 & 0.21\% & (7m)                      \\
\multicolumn{1}{l|}{CPL (8 sols.)}    & 6.13                     & 0.08\%  & \multicolumn{1}{c|}{(33s)} & 10.39 & 0.12\% & \multicolumn{1}{c|}{(3m)}   & 15.71 & 0.16\% & (14m)                     \\ \bottomrule
\end{tabular}
\label{table_TSP_CVRP}
\end{table}

The experimental results on random TSP and CVRP instances with different sizes are shown in Table~\ref{table_TSP_CVRP}. Following the setting in \cite{kool2019attention}, for each problem with different sizes, $10,000$ randomly generated instances are used as the testing data. We compare CPL with (1) an exact solver Concorde~\cite{applegate2006traveling}; (2) two widely-used heuristic solvers LKH3~\cite{helsgaun2000effective,helsgaun2017extension} and OR-Tools~\cite{ortools2019}; (3) five learning-based improvement methods NeuRewriter~\cite{chen2019learning}, Neural 2-Opt~\cite{d2020learning}, Neural Large Neighborhood Search (NLNS)~\cite{hottung2020neural}, Learning Improvement Heuristics in Wu et al.~\cite{wu2021learning}, and DACT~\cite{ma2021learning}; and (4) four constructive neural combinatorial optimization methods Graph Convolutional Network (GCN) with beam search~\cite{joshi2019efficient}, AM~\cite{kool2019attention} with greedy and sampling rollout, MDAM~\cite{xin2021multi} with beam search and POMO~\cite{kwon2020pomo}. Similar to other NCO works, we report the average results (Cost), optimal gap (Gap), and the total run time (Time) for solving all $10,000$ instances. It should be noticed that, even averaging over $10,000$ random instances, the baseline performance could still be slightly different. Therefore we mainly use the gap for comparison as in the most related works~\cite{kwon2020pomo,ma2021learning}.

According to the results in Table~\ref{table_TSP_CVRP}, our proposed CPL method can consistently improve the POMO performance by leveraging the multi-shot prediction on the continuation path. For TSP, it achieves nearly $0\%$ optimality gap to the exact Concorde solver for instances with $20$ and $50$ nodes, and a $0.08\%$ gap for TSP100. These results are comparable with the powerful DACT solver with much faster running time, and outperform other learning-based solvers. For the challenging CVRP, CPL's improvements over POMO are much more significant. The $0.08\%$ to $0.16\%$ optimality gaps to the well-developed LKH3 are promising given the fast inference time and minimal specific domain knowledge required by CPL. 

CPL needs a longer run time than POMO due to the multi-shot prediction, which is the cost for its better performance. However, it still only takes tens of seconds to a few minutes to solve $10,000$ instances on a single GPU, which is much faster than the learning-based improvement method on multiple GPUs~\cite{ma2021learning}. How to design a better continuation path construction and sampling method to further improve the performance-time ratio for CPL could be interesting future work.

\clearpage 
\subsection{Generalization Performance on Realistic TSP Benchmark}

We test our proposed continuation path learning (CPL) method's performance on the TSPLib benchmark problems as shown in Table~\ref{table_TSPLib}. These problem instances have different city distributions and different sizes from $50$ to $200$. We mainly compare CPL with the heuristic solver OR-Tools~\cite{ortools2019}, two improvement-based NCO methods Wu et al.~\cite{wu2021learning} and DACT~\cite{ma2021learning}, and two construction-based NCO methods AM~\cite{kool2019attention} with sampling and POMO~\cite{kwon2020pomo}. All these methods except POMO need to continuously interact with the problem instances (e.g., with $3,000$ to $10,000$ steps) or generate a large number of candidate solutions for selection (e.g., $10,000$ for AM with sampling). For all methods, we follow the setting in DACT~\cite{ma2021learning} to report results on the first five instances (have sizes $51$ to $76$) with models trained on $50$ cities, and the results for the rest instances (have sizes $99$ to $200$) with models trained on $100$ cities.

According to the results, CPL with $8$ continuation solutions can further improve the POMO's already very competitive overall performance from $2.15\%$ to $1.72\%$. This overall result is even better than those for DACT with the strong setting (e.g., with T = $10k$ steps and 4 augmentations). Following DACT~\cite{ma2021learning}, we also report the average optimality gap for instances with different sizes. CPL generally has good performance for instances with sizes from $50$ to $150$, which is close to the training size ($50$ and $100$). However, its performance drastically decreases for instances with larger numbers of cities that are far from the training set. This limitation is shared among other construction-based and improvement-based neural combinatorial optimization methods.

\subsection{Generalization Performance on Realistic CVRP Benchmark}

Table~\ref{table_CVRPLib} shows the results on $22$ realistic CVRPLIB benchmark problems~\cite{uchoa2017new}. They have quite different sizes and depot/nodes distributions to the instances our model has learned (uniformly distributed depot and $100$ nodes), so the generalization ability is important for a good performance. In this case, CPL with $8$ continuation solutions can further improve POMO and achieve a $4.95\%$ average optimality gap. This result is better than the powerful DACT improvement solver with $t=5k$ improvement steps for each instance, but is outperformed by DACT with the strongest setting ($t=10k$ with 6 augmentations). The construction, improvement, and search methods are not necessarily competitors.

\begin{table}[ht]
\setlength{\tabcolsep}{4pt}
\centering
\small
\caption{Experimental results on TSPLib with $50$ to $200$ cities.}
\begin{tabular}{c|ccccc|cc}
\toprule
Instance                                      & \multicolumn{1}{c}{OR-Tools} & \multicolumn{1}{c}{Wu et al.} & \multicolumn{1}{c}{DACT}   & \multicolumn{1}{c}{DACT}    & \multicolumn{1}{c|}{AM-S}    & POMO   & \multicolumn{1}{c}{CPL}   \\
\multicolumn{1}{l|}{}                         & \multicolumn{1}{c}{}         & (T = 3k, M=1k)                & \multicolumn{1}{c}{(T=3k)} & \multicolumn{1}{c}{(T=10k, Aug. $\times 4$)} & \multicolumn{1}{c|}{(T=10K)} &        & \multicolumn{1}{c}{(T=8)} \\ \midrule
eil51                                         & 2.35\%                       & 1.17\%                        & 1.64\%                     & 0.00\%                      & 2.11\%                       & 0.23\% & 0.00\%                    \\
berlin52                                      & 5.34\%                       & 2.57\%                        & 0.03\%                     & 0.03\%                      & 1.67\%                       & 0.00\% & 0.00\%                    \\
st70                                          & 1.19\%                       & 0.89\%                        & 0.44\%                     & 0.30\%                      & 2.22\%                       & 0.00\% & 0.00\%                    \\
eil76                                         & 4.28\%                       & 4.65\%                        & 2.42\%                     & 1.67\%                      & 3.35\%                       & 0.74\% & 0.19\%                    \\
pr76                                          & 2.72\%                       & 1.37\%                        & 1.02\%                     & 0.03\%                      & 2.84\%                       & 0.09\% & 0.02\%                    \\
rat99                                         & 1.73\%                       & 8.51\%                        & 4.05\%                     & 0.74\%                      & 9.50\%                       & 4.62\% & 2.39\%                    \\
KroA100                                       & 0.78\%                       & 2.08\%                        & 0.86\%                     & 0.45\%                      & 79.49\%                      & 1.05\% & 0.68\%                    \\
KroB100                                       & 3.91\%                       & 5.78\%                        & 0.27\%                     & 0.25\%                      & 9.30\%                       & 0.78\% & 0.73\%                    \\
KroC100                                       & 4.02\%                       & 3.17\%                        & 1.06\%                     & 0.84\%                      & 8.04\%                       & 0.38\% & 0.17\%                    \\
KroD100                                       & 1.61\%                       & 5.00\%                        & 3.54\%                     & 0.12\%                      & 10.02\%                      & 1.92\% & 1.25\%                    \\
KroE100                                       & 2.40\%                       & 3.29\%                        & 2.17\%                     & 0.32\%                      & 3.10\%                       & 1.39\% & 1.02\%                    \\
rd100                                         & 3.53\%                       & 0.06\%                        & 0.08\%                     & 0.00\%                      & 1.93\%                       & 0.00\% & 0.09\%                    \\
eil101                                        & 5.56\%                       & 4.61\%                        & 3.66\%                     & 2.86\%                      & 3.97\%                       & 0.16\% & 0.00\%                    \\
lin105                                        & 3.09\%                       & 2.48\%                        & 3.41\%                     & 0.69\%                      & 32.13\%                      & 0.77\% & 0.70\%                    \\
pr107                                         & 1.74\%                       & 3.87\%                        & 5.86\%                     & 3.81\%                      & 43.26\%                      & 1.35\% & 0.90\%                    \\
pr124                                         & 5.91\%                       & 2.97\%                        & 1.56\%                     & 1.22\%                      & 4.41\%                       & 0.08\% & 0.08\%                    \\
bier127                                       & 3.76\%                       & 3.48\%                        & 4.08\%                     & 2.46\%                      & 1.71\%                       & 4.31\% & 5.00\%                    \\
ch130                                         & 2.85\%                       & 4.89\%                        & 6.63\%                     & 1.93\%                      & 2.96\%                       & 0.10\% & 0.02\%                    \\
pr136                                         & 5.62\%                       & 6.33\%                        & 5.54\%                     & 4.54\%                      & 4.90\%                       & 0.74\% & 0.93\%                    \\
pr144                                         & 1.28\%                       & 1.40\%                        & 3.44\%                     & 2.49\%                      & 8.77\%                       & 0.50\% & 0.66\%                    \\
ch150                                         & 3.08\%                       & 3.55\%                        & 3.60\%                     & 1.23\%                      & 3.45\%                       & 0.44\% & 0.41\%                    \\
KroA150                                       & 4.03\%                       & 4.51\%                        & 6.93\%                     & 3.91\%                      & 9.98\%                       & 0.79\% & 0.69\%                    \\
KroB150                                       & 5.52\%                       & 5.40\%                        & 6.10\%                     & 2.82\%                      & 9.87\%                       & 1.94\% & 1.49\%                    \\
pr152                                         & 2.92\%                       & 2.17\%                        & 4.48\%                     & 3.59\%                      & 13.47\%                      & 1.23\% & 0.87\%                    \\
u159                                          & 8.79\%                       & 7.67\%                        & 6.84\%                     & 3.16\%                      & 7.38\%                       & 1.00\% & 0.97\%                    \\
rat195                                        & 2.84\%                       & 9.90\%                        & 6.93\%                     & 4.99\%                      & 16.57\%                      & 9.60\% & 6.50\%                    \\
d198                                          & 1.16\%                       & 4.99\%                        & 12.27\%                    & 8.75\%                      & 331.58\%                     & 21.9\% & 20.1\%                    \\
KroA200                                       & 1.27\%                       & 7.01\%                        & 3.60\%                     & 1.25\%                      & 15.64\%                      & 2.05\% & 1.78\%                    \\
KroB200                                       & 3.67\%                       & 7.05\%                        & 10.51\%                    & 5.66\%                      & 18.54\%                      & 4.25\% & 2.29\%                    \\ \midrule
\multicolumn{1}{l|}{Avg. Gap for {[}50,100)}  & 2.93\%                       & 3.19\%                        & 1.60\%                     & 0.46\%                      & 3.61\%                       & 0.95\% & 0.43\%                    \\
\multicolumn{1}{l|}{Avg. Gap for {[}100,150)} & 3.29\%                       & 3.53\%                        & 3.01\%                     & 1.57\%                      & 15.29\%                      & 0.97\% & 0.87\%                    \\
\multicolumn{1}{l|}{Avg. Gap for (150,200{]}} & 3.70\%                       & 5.81\%                        & 6.81\%                     & 3.93\%                      & 47.39\%                      & 4.80\% & 3.89\%                    \\
\multicolumn{1}{l|}{Avg. Gap for All}         & 3.34\%                       & 4.17\%                        & 3.90\%                     & 2.07\%                      & 22.83\%                      & 2.15\% & 1.72\%                    \\ \bottomrule
\end{tabular}
\label{table_TSPLib}
\vspace{-0.2in}
\end{table}

\clearpage

\begin{table}[ht]
\centering
\begin{threeparttable}
\caption{Experimental results on CVRPLIB instances with different sizes and distributions}
\begin{tabular}{cc|cccc|cc}
\toprule
\multicolumn{1}{c|}{Instance}   & (Depot, Nodes) & OR-Tools & DACT    & DACT    & AM-S    & POMO   & CPL    \\
\multicolumn{1}{c|}{}           & Type           &          & (T=5k)  & (T=10k, 6 aug.) & (T=10K) &        & (T=8)  \\ \midrule
\multicolumn{1}{c|}{X-n101-k25} & (R, R)         & 6.57\%   & 2.09\%  & 1.47\%  & 32.95\% & 6.73\% & 3.95\% \\
\multicolumn{1}{c|}{X-n106-k14} & (E, C)         & 3.72\%   & 2.93\%  & 1.87\%  & 6.78\%  & 2.27\% & 2.15\% \\
\multicolumn{1}{c|}{X-n110-k13} & (C, R)         & 7.87\%   & 1.43\%  & 0.13\%  & 3.15\%  & 1.22\% & 0.89\% \\
\multicolumn{1}{c|}{X-n115-k10} & (C, R)         & 4.50\%   & 3.29\%  & 1.68\%  & 7.52\%  & 9.96\% & 4.33\% \\
\multicolumn{1}{c|}{X-n120-k6}  & (E, RC)        & 6.83\%   & 3.50\%  & 2.38\%  & 4.54\%  & 9.32\% & 8.47\% \\
\multicolumn{1}{c|}{X-n125-k30} & (R, C)         & 5.63\%   & 6.51\%  & 5.44\%  & 35.16\% & 5.78\% & 4.17\% \\
\multicolumn{1}{c|}{X-n129-k18} & (E, RC)        & 8.37\%   & 2.93\%  & 2.55\%  & 4.00\%  & 2.16\% & 1.02\% \\
\multicolumn{1}{c|}{X-n134-k13} & (R, C)         & 21.61\%  & 6.98\%  & 2.63\%  & 20.13\% & 3.98\% & 2.29\% \\
\multicolumn{1}{c|}{X-n139-k10} & (C, R)         & 12.02\%  & 2.54\%  & 2.08\%  & 4.30\%  & 3.24\% & 2.63\% \\
\multicolumn{1}{c|}{X-n143-k7}  & (E, R)         & 11.27\%  & 7.80\%  & 3.55\%  & 8.88\%  & 4.34\% & 2.72\% \\
\multicolumn{1}{c|}{X-n148-k46} & (R, RC)        & 7.80\%   & 2.69\%  & 2.22\%  & 79.53\% & 9.78\% & 4.73\% \\
\multicolumn{1}{c|}{X-n153-k22} & (C, C)         & 8.01\%   & 11.04\% & 6.53\%  & 78.11\% & 14.4\% & 11.4\% \\
\multicolumn{1}{c|}{X-n157-k13} & (R, C)         & 2.57\%   & 4.64\%  & 3.12\%  & 16.30\% & 8.66\% & 7.76\% \\
\multicolumn{1}{c|}{X-n162-k11} & (C, RC)        & 6.31\%   & 4.43\%  & 2.62\%  & 6.37\%  & 6.00\% & 5.93\% \\
\multicolumn{1}{c|}{X-n167-k10} & (E, R)         & 9.34\%   & 5.37\%  & 3.47\%  & 8.41\%  & 3.61\% & 3.08\% \\
\multicolumn{1}{c|}{X-n172-k51} & (C, RC)        & 10.74\%  & 6.23\%  & 3.41\%  & 85.37\% & 10.7\% & 7.01\% \\
\multicolumn{1}{c|}{X-n176-k26} & (E, R)         & 8.99\%   & 10.29\% & 5.93\%  & 20.39\% & 10.6\% & 6.84\% \\
\multicolumn{1}{c|}{X-n181-k23} & (R, C)         & 2.94\%   & 3.41\%  & 2.08\%  & 6.45\%  & 5.57\% & 4.42\% \\
\multicolumn{1}{c|}{X-n186-k15} & (R, R)         & 7.75\%   & 5.99\%  & 4.94\%  & 6.01\%  & 6.58\% & 6.03\% \\
\multicolumn{1}{c|}{X-n190-k8}  & (E, C)         & 6.53\%   & 7.97\%  & 6.73\%  & 46.61\% & 6.68\% & 6.15\% \\
\multicolumn{1}{c|}{X-n195-k51} & (C, RC)        & 13.76\%  & 7.00\%  & 4.36\%  & 79.26\% & 13.7\% & 7.98\% \\
\multicolumn{1}{c|}{X-n200-k36} & (R, C)         & 4.15\%   & 5.93\%  & 5.89\%  & 26.25\% & 5.97\% & 4.87\% \\ \midrule
\multicolumn{2}{l|}{Average Optimality Gap}      & 8.06\%   & 5.23\%  & 3.41\%  & 26.66\% & 6.88\% & 4.95\% \\ \bottomrule
\end{tabular}
\begin{tablenotes}\footnotesize
\item[*] Three types of depot positions: C-Central, E-Eccenric/Corner, R-Random.
\item[*] Three types of node distributions: R-Random, C-Clustered, RC-Mixed Random and Clustered.
\end{tablenotes}
\label{table_CVRPLib}
\end{threeparttable}
\end{table}


\end{document}